\documentclass[twoside,11pt]{article}
\usepackage{jmlr2e}
\usepackage{natbib}
\usepackage[utf8]{inputenc}
\usepackage{lineno}
\usepackage{amsmath}
\usepackage{mathrsfs}
\usepackage{amsfonts}
\usepackage{dsfont}
\usepackage{booktabs}
\usepackage{placeins}
\usepackage{float}
\usepackage[noend,ruled,vlined]{algorithm2e}
\usepackage[noend]{algpseudocode}
\usepackage{appendix}
\usepackage{framed}\usepackage{multicol}
\usepackage{nomencl}

\usepackage{etoolbox}
\renewcommand\nomgroup[1]{%
	\item[]\hspace*{-\leftmargin}\newline%
	\item[\bfseries
	\ifstrequal{#1}{A}{Acronyms}{%
		\ifstrequal{#1}{B}{Variables and Functions}{%
			\ifstrequal{#1}{C}{Parameters}{%
				\ifstrequal{#1}{D}{Sets}{%
	}}}}%
	]\hspace*{-\leftmargin}\newline%
}

\algrenewcommand\algorithmiccomment[2][\small]{{\hfill#1\quad\qquad\qquad\(\triangleright\) \texttt{#2}}}
\newcommand{\argmin}[1]{\underset{#1}{\text{argmin}}\,}

\usepackage{float}
\restylefloat{table}

\ShortHeadings{}{Nespoli and Medici}
\firstpageno{1}

\begin{document}

\title{Multivariate Boosted Trees and Applications to Forecasting and Control}

\author{\name Lorenzo Nespoli \email lorenzo.nespoli@supsi.ch\\
       \addr ISAAC, SUPSI\\
       Mendrisio, CH
       \AND
       \name Vasco Medici \email vasco.medici@supsi.ch\\
       \addr ISAAC, SUPSI\\
       Mendrisio, CH}

\editor{arXiv}

\maketitle
\makenomenclature
\begin{abstract}
Gradient boosted trees are competition-winning, general-purpose, non-parametric regressors, which exploit sequential model fitting and gradient descent to minimize a specific loss function. The most popular implementations are tailored to univariate regression and classification tasks, precluding the possibility of capturing multivariate target cross-correlations and applying structured penalties to the predictions.  
In this paper, we present a computationally efficient algorithm for fitting multivariate boosted trees. We show that multivariate trees can outperform their univariate counterpart when the predictions are correlated. Furthermore, the algorithm allows to arbitrarily regularize the predictions, so that properties like smoothness, consistency and functional relations can be enforced. We present applications and numerical results related to forecasting and control. 
\end{abstract}


\begin{table*}
	\begin{framed}
		\begin{multicols}{2}	
			\printnomenclature
		\end{multicols}
	\end{framed}
\end{table*}

\section{Introduction}
We propose the use of multivariate boosted trees (MBTs) to induce arbitrary regularization and consistency properties in the tree output. This can be done both via penalization of the multivariate output or requiring it to be a superposition of basis functions. Inducing regularization in multivariate output is not new, but while this is common for example in neural network architectures  \citep{Oreshkin2019,Belharbi2018,Bronstein2017}, they are currently not exploited in tree-based algorithms. One exception is the possibility of LightGBM and XGBoost to express monotonicity conditions of the univariate prediction, with respect to a given input \citep{lightGBMmanual}. This is obtained by inhibiting the tree growth if the new leaf causes a non-monotonic split in the selected feature. However, this may produce unnecessarily shallow trees if not enough split candidates are tested, which could be the case if the tree is grown using histogram search, one of the most popular methods for finding candidate splits.

\subsection{Related work}
In \cite{Pande2017}, an MBT tailored to predicting longitudinal data is presented. This kind of data is typically generated in medical studies, sampling the population at different points in time. Typically, the amount of available data to model the temporal relation is limited. The authors developed MBTs and trained them in function space, using B-Splines to model time interactions. The algorithm is tested on a synthetic dataset, generated using simple algebraic formulae to model the target dependence over features and time. In \cite{Li2019}, a single tree is fitted using a multivariate linear regressor as weak learner. The tree is grown such that in each leaf the dataset is divided into two classes, based on the points for which the tree returned an overshot or undershot prediction. Despite the interesting idea, splitting points are not chosen with a variance reduction criterion, and only one model is fitted, thus not exploiting gradient boosting. The algorithm is found to perform better than linear regression on 3 machine learning datasets, while performance against LightGBM is datasets dependent. Recently, the authors in \cite{Zhang2019} proposed a multivariate version of the XGBoost algorithm, introduced a new histogram algorithm for datasets with sparse features and implemented a performance tailored C++ library. In this work, we make use of the same approach to fit MBTs, coupling it with non-constant response functions.   

\subsection{Contributions}
We have extended the formulation of boosted trees to the multivariate and non-constant response cases. This goes beyond popular gradient boosting libraries, which adopt a univariate and constant response paradigm. To the best of our knowledge, no one has ever presented a non-constant response MBT. This new method allows us to arbitrarily regularize the covariance structure of the outputs and induce smoothness, which are relevant features for many applications.\\
In section \ref{sec:quantile_th}, we introduce a smoothed formulation of the quantile loss and show its superiority in terms of expected quantile loss and crossings of the predicted quantile. In section \ref{sec:hierarchical_th}, we introduce a new approach for hierarchical forecasting, which takes into account previous forecast error, and show that this method is better compared to other state-of-the-art algorithms for the first prediction steps. This is possible thanks to the introduction of a consistent non-constant response function. Finally, in section \ref{sec:vsc_th}, as an example of application, we present a way to fit voltage sensitivity coefficients for electrical distribution networks through boosted trees, while retaining their linear form w.r.t. the active and reactive powers. The fit is based on few exogenous variables, and we show that robustness to input variable noise makes this approach suitable for control application.  
The algorithm has been released as a python package under MIT license, and it is freely available at \url{https://github.com/supsi-dacd-isaac/mbtr}. All the code used for running the experiments presented in the paper, including the code for generating the figures, is available at \url{https://github.com/supsi-dacd-isaac/mbtr_experiments}. All the used datasets are freely accessible, and directly downloaded by the experiment's code. The dataset used for the numerical experiments can be downloaded from \url{https://zenodo.org/record/4108561#.YEeukVmYWV5} and \url{https://zenodo.org/record/4549296#.YEeuvFmYWV4}.

\section{Background}\label{sec:background}
Given a matrix of targets $y \in \mathds{R}^{N \times n_t}$, where $N$ is the number of observations and $n_t$ the dimensionality of the target, and a set of features (or covariates, or explanatory variables) $x \in \mathds{R}^{N \times n_f}$, we call the union of their observations a dataset $\mathcal{D} = \{(x_i,y_i)_{i=1}^N\}$. Our goal is to fit a learnable model $F(x, \Theta)$, where $\Theta$ is the set of model's parameters, on dataset $\mathcal{D}$, such that it minimizes the expected loss on unseen data. To achieve this, we minimize the empirical expectation of the loss function $\ell(y_i, F(x_i, \Theta)): \mathds{R}^{n_t} \rightarrow \mathds{R}$, also known as empirical risk, on the observed dataset $\mathcal{D}$:
\begin{equation}
\Theta^* = \argmin{\Theta} \sum_{i=1}^N \ell(y_i, F(x_i, \Theta))
 = \argmin{\Theta} \mathscr{L}(y, F(x, \Theta))
\end{equation}

\subsection{Decision trees}\label{sec:trees}
Since GBTs use regression trees as weak learners, we recall here their formal description and fitting strategy. A regression tree is a function partitioning the input space $\mathds{R}^{n_f}$ into different regions, or leaves, each of which contains a response function $r(w_l): \mathds{R}^{n_w}\rightarrow\mathds{R}^{n_t}$, $n_t=1$ corresponding to a univariate tree, parametrized by weights $w_l \in \mathds{R}^{n_w}$. 
Formally, a tree can be described as a function $f(x, \theta) \in \mathcal{F}$ where $\mathcal{F}=\left\{f(x, \theta)=r(w_{q(x)})\right\}\left(q: \mathbb{R}^{n_f} \rightarrow n_l\right)$ where $q$ represents the structure of the tree which maps observations into leaf indexes and $\theta$ is the set of the tree's parameters. Equivalently, a tree can be described as the sum of the leaves' response functions, weighted by the indicator function $\mathds{1}_{l}(x_i)=\left\{i \mid q\left(x_{i}\right)=l\right\}: \mathds{R}^{n_f} \rightarrow \{0,1\}$, returning 1 if $x_i$ belongs to the $l_{th}$ leaf, and 0 otherwise:
\begin{linenomath*}\begin{equation}\label{eq:tree}
	f(x_i,\theta) = \sum_{l=1}^{n_l} r(w_l) \mathds{1}_l(x_i)
	\end{equation}\end{linenomath*}
In this paper we will only consider trees applying a recursive binary partitioning (or splits) of the input to construct their leaves, resulting in leaves that are disjoint and orthogonal w.r.t. the features under consideration. In this case, $\theta = \{S,W\}$ consists of the ordered set of variables and levels defining the splits for each of the $n_n$ nodes of the tree, $S = \{(v_n, l_n)_{n=1}^{n_n} \}$, and the parameter set of the response functions for each leaf of the tree, $W=\{w_l\}_{l=1}^{n_l}$. While in this paper we will make use of different response functions, in the standard case this is a constant, thus $r(w_l) = w_l$, $w_l \in \mathds{R}^{n_w}$. In order to fit both univariate and multivariate trees, we can rely on the following remark:
\\ \\
{\bf Remark}{\label{rem:1}
	\it
	Since the functional form $r(w_l)$ is the same for each leaf, $w_l$ is constant for a given leaf, and since the leaves are disjoint regions of the feature space, we only need to know the functional form of the leaves' loss function in order to fit a tree.
}
\\ \\
We can then write the total loss, as a summation of the leaf losses: 
\begin{align}
\mathscr{L}(y, f(x, \theta)) &=\sum_{i=1}^{N} \ell\left(y_i, f\left(x_i, \theta\right)\right) \nonumber \\
&=\sum_{i=1}^{N} \ell\left(y_{i}, \sum_{l=1}^{L} r\left(w_{l}\right) \mathds{1}_{l}(x_i)\right) \nonumber \\
&=\sum_{l=1}^{n_l}\sum_{i \in \mathcal{I}_l}  \ell\left(y_{i}, r\left(w_{l}\right)\right) = \sum_{l=1}^{n_l} \ell_l \label{eq:leaf_loss}
\end{align}
where $\mathcal{I}_l = \{i: \mathds{1}_l(x_i) = 1 \}$. 
To fit the tree, we must find both the optimal values of $w_l$ inside a given leaf, and the leaf partitions $\mathcal{I}_l$. While the first task is straightforward, the second one is much harder; in fact, since the latter is usually computationally infeasible, greedy algorithms are used to find the best splits. Basically, at each iteration, a leaf with dataset $\mathcal{D}_l$ is split if the sum of the loss computed on the partial datasets $\mathcal{D}_{l,s1}$ and $\mathcal{D}_{l,s2}$ is lower than the leaf loss. It is easy to see that the splitting criterion (that is, how to divide $\mathcal{D}_l$), must be only dependent on the features $x$ since at prediction time we won't know the values of $y$. Even if this approach is simple, it can result in high computational costs; in the extreme case in which all the points are regarded as splitting candidates, the computational cost of the algorithm is $\mathcal{O}(N \times n_f)$ for the first splitting decision. In this paper, we restrict splitting candidates using histograms, as done in LightGBM \citep{Ke2017}. This reduces the cost of finding the optimal split to $\mathcal{O}(n_{qs} \times n_f)$ where $n_{qs}$ is the number of considered bins. Note that if conditions stated in the remark were not met, it would be harder to optimize the tree's parameters $\theta$. If the reward function was not the same in all the leaves, we should decide which response to use in each leaf, based on some optimization strategy. If the leaves were not disjoint, we would end up with overlapping sets $\mathcal{I}_l$, which would be harder to optimize even using greedy algorithms. Finally, having non constant $w_l$ in a given leaf would be equivalent to have a tree with further splits.

\subsection{Boosted trees}\label{sec:gradient_boosting}
Boosting algorithms have progressively gained popularity among the machine learning and statistics community, starting from the introduction in the 90s of the famous AdaBoost classification algorithm \citep{Freund1997}. Originally introduced as an ensemble method \citep{Buhlmann2007}, boosting was later interpreted as a gradient descent in function space \citep{Breiman1998}, opening up the possibility of using it for optimizing a wide variety of smooth and non-smooth objective functions. In this paper, we follow the interpretation of boosting as an iterative optimization strategy for statistical learning. In this section, we review the original gradient descent interpretation in function space presented in \cite{Friedman2001}.
A boosted tree can be described as an additive model of $K$ weak learners, each of which is a tree:
\begin{equation}
F_K(x) = \sum_{k=1}^K f(x, \theta_k), \quad f_k(x, \theta_k) \in \mathcal{F} 
\end{equation}
Under the hypothesis that $\mathscr{L}(y,F)$ is continuous and smooth almost everywhere, we can seek its minimizer $F^*$ through gradient descent iterations. To simplify the notation, we refer to $\frac{\partial \ell(y, F_{k}(x, \Theta))}{\partial F_{k}(x, \Theta)}$ as $g_{k}$, that is, the gradient of the loss with respect to the model's predictions at iteration $k$. As it is known, applying gradient descent to $\mathscr{L}(y, F_{k}(x, \Theta))$ in the $F_{k}(x, \Theta)$ argument is equivalent to solve the following minimization problem (where $F_{k, i} = F_{k}(x_i, \Theta)$ for sake of notation) at each iteration $k$:

\begin{linenomath*}\begin{align}\label{eq:gradient_descent}
	F_{k+1} &= \argmin{F} \mathscr{L}(y, F_{k})+ \frac{\partial \mathscr{L}(y, F_{k})}{\partial F_{k}}^T\left(F -F_{k}\right) + \frac{1}{2\rho} \Vert F - F_k\Vert_2^2 \\
	&= \argmin{F} \mathscr{L}(y, F_{k})+ \sum_{i=1}^Ng_i(F_i-F_{k, i})+ \frac{1}{2\rho} \Vert F - F_k\Vert_2^2
	\end{align}\end{linenomath*}
where $\Vert \cdot \Vert_2^2$ denotes the sum of squares over all the predictions, $\rho$ is a hyper-parameter and the last equality holds under the assumption of sufficient regularity, so that one can interchange differentiation and integration. Equation \eqref{eq:gradient_descent} can be interpreted as the act of minimizing the first order approximation of the loss function in its argument $F$, while trying not to deviate too much from the predictions of the previous fitted model $F_{k}$. In order to find the minimizer of \eqref{eq:gradient_descent}, we apply the first order optimality condition, w.r.t. each observation, and we find:
\begin{linenomath*}\begin{equation}\label{eq:gradient_minimizer}
	F_{k+1}(x, \Theta) = F_{k}(x, \Theta) - \rho g_{k}
	\end{equation}\end{linenomath*}
which is the gradient descent step. The loss gradient $g_k$ is easily computed for the dataset $\mathcal{D}$. However, as pointed out in \cite{Friedman2001}, our goal is to minimize $\mathscr{L}(y, F)$ not only for the dataset $\mathcal{D}$, but also on unseen data, in order to perform statistical learning and achieve model generalization. For this reason, boosting replaces $g_k$ with the gradient \emph{learned} by a base model $f(x,\theta)$, also known as weak learner.     
The iterative model fitting becomes:
\begin{linenomath*}\begin{equation}\label{eq:additive_model}
	F_{k+1}(x, \Theta) =  F_{k}(x, \Theta) - \rho f_{k}(x, \theta)
	\end{equation}\end{linenomath*}
where $f_{k}(x,\theta)$ has been fitted under least squares criterion on $g_{k}$. Boosting in function space is a building block of many other machine learning algorithms. For example, it has been recently adopted, in combination with parametric probabilistic modelling and the concept of natural gradient, in the NGBoost library \citep{Duan2019}.
In this paper, we will follow the method adopted by XGboost and LightGBM, which optimizes the boosted tree using a second-order approximation of the loss function. We retain only the additive stage-wise strategy defined by the iteration \eqref{eq:additive_model}, assuming it to be coercive with respect to the prediction error. Indeed the presence of the learning rate $\rho$ helps in dampening the response of the current iteration model, avoiding overshooting of the final model $F_{k+1}$. Under a stage-wise strategy, we can write the second order approximation of $\mathscr{L}(y, F_{k})$ with respect to the new weak-learner as:
\begin{equation}\label{eq:loss_approx}
\mathscr{L}(y, F)  \simeq  \mathscr{L}(y, F_{k})+ \sum_{i=1}^N g_{k, i} f(x_i, \theta) + \frac{1}{2} h_{k,i} f^2(x_i, \theta)+ \frac{\lambda }{2} \sum_{l=1}^{n_l}w_l^2
\end{equation}
where $h = \frac{\partial^2 \ell(y, F_{k}(x, \Theta))}{\partial F(x, \Theta)^2}$ is the second order derivative of the loss w.r.t. the predictions and the last term is a regularization term. At each stage we want to find the optimal set of parameters $\theta_k^*$ which includes both the split points and the weights. To find $\theta_k^*$ we can follow the same strategy to fit a tree introduced in section \ref{sec:trees}, using the second order approximation of the loss function. At first, \eqref{eq:loss_approx} is used to estimate the loss in each leaf, given the current splits $\{\mathds{1}_l \}_1^{n_l}$, and secondly, a greedy strategy is applied to find the optimal splits.
In the case in which the model response is constant in each leaf, and equal to $r(w_l) = w_l$, we can rewrite \eqref{eq:leaf_loss} using the second order approximation \eqref{eq:loss_approx}; the loss function (disregarding the constant term) can be defined as summation of leaf losses:
\begin{linenomath*}\begin{equation}\label{eq:approximate_loss}
	\mathscr{L}(y, f_{k})  \simeq \sum_{l=1}^{n_l}\left[ \sum_{i \in \mathcal{I}_l}  \left( g_{k, i} w_l + \frac{1}{2} h_{k, i} w_l^2\right) + \frac{\lambda}{2}  w_l^2\right]
\end{equation}\end{linenomath*}
Thus, for the $l_{th}$ leaf, the optimal $w_l$ given the split is:
\begin{linenomath*}\begin{equation}\label{eq:optimal_response}
	w_l^* = \frac{- \sum_{i \in \mathcal{I}_l} g_{k, i}}{\lambda + \sum_{i \in \mathcal{I}_l}h_{k, i}}
	\end{equation}\end{linenomath*}
The optimal approximated leaf loss becomes:
\begin{linenomath*}\begin{equation}\label{eq:optimal_loss}
	\tilde{\ell}^* = -\frac{1}{2}\frac{\sum_{i\in\mathcal{I}_l}g_{k, i}^2}{\lambda + \sum_{i\in\mathcal{I}_l} h_{k,i}}
	\end{equation}\end{linenomath*}

This is the same procedure used by XGboost an LightGBM, for instance. 
In order to consider non-constant responses, two strategies can be followed: the first is to replace $w_l$ in the inner summation of \eqref{eq:approximate_loss} with $r(w_l)$. We can then compute the optimal response as:

\begin{linenomath*}\begin{equation}\label{eq:optimal_response_2}
	r(w_l)^* = \frac{- \sum_{i \in \mathcal{I}_l} g_{k, i}}{\lambda + \sum_{i \in \mathcal{I}_l}h_{k, i}}
\end{equation}\end{linenomath*}
In order to find the optimal parameters $w_l^*$, this requires the response $r(w_l)$ to be analytically known and invertible. Since this is not true for some interesting applications, as in the case in which the response is in the form $r(w_l) = A w$ with $\mathds{R}^{n_a\times n_t}$ and $n_a > n_t$, we propose to replace the approximation of the loss function w.r.t. the model's prediction with the approximation w.r.t. the models' weights $w$. Defining $\tilde{g}$ and $\tilde{h}$ as the gradient and the second derivative of the loss function, with respect to the model \emph{weights}, for the chain rule, we can write for each leaf:
\begin{align}\label{eq:gradient_hessian}
\tilde{g}_{k, i} &= g_{k, i} \frac{\partial r(w_l)}{\partial w_l} &\tilde{h}_{k, i} &= g_{k, i} \frac{\partial^2 r(w_l)}{\partial w_l^2} + h_{k, i} \left(\frac{\partial r(w_l)}{\partial w_l }\right)^2 \qquad \forall i\in\mathcal{I}_l
\end{align}
Note that for the usual case in which the leaf response is constant, $\tilde{g} = g$ and $\tilde{h} = h$. We can now use equations \eqref{eq:approximate_loss}, \eqref{eq:optimal_response}  and \eqref{eq:optimal_loss} replacing $g_{k, i}$ and $h_{k, i}$ with $\tilde{g}_{k, i}$ and $\tilde{h}_{k, i}$. This allows us to keep the same procedure for fitting the tree while just requiring $r(w)$ to be differentiable w.r.t. $w$.

\subsection{MBTs}\label{sec:multivariate_gbt}
Multivariate GBTs can be fitted by following the same procedure described in the previous section. The only difference relies on the dimensionality of the target variable $y \in \mathds{R}^{N\times n_t}$ where $n_t$ is strictly greater than 1, and the use of the Hessian matrix instead of the second derivative for the the computation of the approximated loss and optimal weights. For clarity, we report the matrix form of $\tilde{g}$ and $\tilde{h}$ in the multivariate case, for which \eqref{eq:gradient_hessian} are the univariate analogous:
\begin{align}
\tilde{g} &= g \left(\frac{\partial r(w_l)}{\partial w_l}\right)\label{eq:gradient}\\
\tilde{h} &= g \frac{\partial^2 r(w_l)}{\partial w_l^2} + \frac{\partial r(w_l)}{\partial w_l } ^T h \frac{\partial r(w_l)}{\partial w_l }\label{eq:hessian}
\end{align}
where $g \in \mathds{R}^{N\times n_t}$, $h \in \mathds{R}^{N\times n_t\times n_t}$, $\frac{\partial r(w_l)}{\partial w_l} \in \mathds{R}^{n_t\times n_w}$, $\frac{\partial^2 r(w_l)}{\partial w_l^2} \in \mathds{R}^{n_t \times n_w \times n_w}$. Note that the number of dimensions of the leaf parameter vector, $n_w$, may be different from the dimensionality of the target, $n_t$. For example, this is the case of hierarchical forecasting, presented in section \ref{sec:hierarchical_th}. We stress out that in the multivariate case, the second derivative of the response function is a 3-order tensor. However, as we will see, for many combinations of objective function and responses, MBT fitting won't require to store or compute the whole tensor, considerably simplifying the computational effort.
For the sake of notation, replacing $\sum_{i \in \mathcal{I}_l} \tilde{g}_{i,k}$ with $\tilde{G}$ and $\sum_{i\in\mathcal{I}_l} \tilde{h}_{i,k}$ with $\tilde{H}$, the optimal response \eqref{eq:optimal_response} and the optimal loss \eqref{eq:optimal_loss} can be rewritten as:

\begin{align}
w_l^* &= -\left(\Lambda + \tilde{H} \right)^{-1} \tilde{G} \label{eq:optimal_response_mgb} \\
\mathscr{L}_l^* &= \tilde{G}^T\left(\Lambda + \tilde{H} \right)^{-1} \tilde{G} \label{eq:optimal_loss_mgb}
\end{align}
where $\Lambda \in \mathds{R}^{n_r \times n_r}$ is the quadratic regularization matrix, which weights the L2 norm penalization of the model parameters, $\Vert w_l \Vert_{\Lambda}^2$.   
The complete procedure for fitting the MBT is described in algorithm \ref{alg:mgbt} and \ref{alg:fit_tree}. Algorithm $\ref{alg:mgbt}$ describes the boosting procedure: starting from an initial guess for $\hat{y}$, which in this case corresponds to the column-expectations of $y$, we retrieve the gradient $\tilde{g}$ and hessian matrices $\tilde{h}$ for all the observations of the dataset (line 3), given the loss function $\mathscr{L}$ and the leaf response function $r$.
At line 4 the weak learner at iteration $k$ is fitted using the $\textbf{\texttt{fit-tree}}$ algorithm described in \ref{alg:fit_tree}. Then the overall model $F_k$ is updated (line 5) along with the training loss (line 7). This is computed through the exact formulation of the loss function and includes a term for the penalization of the number of leaves $T = \sum_{k=1}^K n_{l, k}$ in the final model $F_k$:
\begin{equation}\label{eq:total_loss}
	\varepsilon = \mathscr{L}(y,F_k(x)) + \rho_T T
\end{equation}
The procedure ends if the training loss is not decreasing or the iterations exceeded the maximum number $n_i$. Algorithm \ref{alg:fit_tree} describes the recursive procedure to fit the multivariate tree. At line 1-2 the algorithm halts if the number of observations is lower than a threshold, $n_{min}$. If this is not the case, the total leaf loss is computed (line 3), and the best split point search is carried out for all the variables in $x$ (line 4). As anticipated, we use the same histogram search adopted in XGboost and LightGBT, see algorithm 2 in \cite{Chen2016} and algorithm 1 in \cite{Ke2017}. Briefly speaking, instead of enumerating all the possible split points as done by the pre-sorting algorithm \citep{Mehta1996}, only a few numbers of quantiles are tested for each feature. This does not reduce too much the final regressor accuracy; on the other hand, since finding the best split takes most of the computational time of boosted tree algorithms, this procedure substantially speeds up the fitting process. At line 5, the quantiles for the $j_{th}$ feature are retrieved and are then used at line 7 to obtain the partial sums of the gradient and Hessian, based on the split point $q$ and variable $j$. The split-loss $\mathscr{L}_a + \mathscr{L}_b$ is then computed using equation \eqref{eq:optimal_loss_mgb}; if this value is lower than the current minimum, the latter and the best split candidate are updated (line 10-11). Finally, if a split with a total loss lower than $\mathscr{L}_0$ has been found, the procedure is called recursively, with partial datasets, gradients and Hessian, based on the best split. Otherwise, the current node is considered a terminal leaf, and the optimal response is computed based on equation \eqref{eq:optimal_response_mgb}.

\begin{algorithm}
	\SetAlgoNoLine
	\LinesNumbered
	\DontPrintSemicolon
	\SetNlSty{texttt}{}{}
	\KwIn{training dataset: $\mathcal{D}_{tr} = \{(x_i,y_i)_{i=1}^N\}, \mathscr{L}, n_i, r(w)$}  
	\KwOut{boosted tree $F$}
	$\hat{y} \leftarrow [\mathds{E}_j y_{j,i}]_{i=1}^{n_t}$  \Comment{initial guess} \;
	\While{$k<n_i$ and $\varepsilon_k<\varepsilon_{k-1}$ $, k \scriptscriptstyle+\scriptscriptstyle+$}{
		$\tilde{g}, \tilde{h} \leftarrow \hat{y},y, \mathscr{L}$ \Comment{using \eqref{eq:gradient} and \eqref{eq:hessian}}\;
		$f_k \leftarrow$  $\textbf{\texttt{fit-tree}}\left(x,y,\tilde{g}, \tilde{h}\right)$\;
		$F_k \leftarrow F_{k-1} + \rho f_k $\;
		$\hat{y}\leftarrow F_k(x)$\;	
		$\varepsilon_{k-1} \leftarrow y, F_k(x)$ \Comment{using \eqref{eq:total_loss}}	\;
	}
	\caption{MBT training}\label{alg:mgbt}
\end{algorithm}

\begin{algorithm}
	\SetAlgoNoLine 
	\LinesNumbered
	\DontPrintSemicolon
	\SetNlSty{texttt}{}{}
	\KwIn{$x,y,\tilde{g}, \tilde{h}, f, node$}
	\KwOut{tree $f$}
	\If{$length(x) < n_{min}$}{\KwRet{}}
	$\mathscr{L}_0 = \mathscr{L}^*\leftarrow G,H$ \;
	\For(\Comment{find best split}){$x_j \in x^T$}{ 
		$q_j \leftarrow x_j, n_{qs}$ \;
		\For(\Comment{histogram search}){$q \in q_j$}{ 
			$\tilde{G}_{a},\tilde{H}_{a}, \tilde{G}_{b},\tilde{H}_{b} \leftarrow \tilde{g}, \tilde{h}, q, j $\; 
			$\mathscr{L}_{a},\mathscr{L}_{b}\leftarrow \tilde{G}_{a},\tilde{H}_{a}, \tilde{G}_{b},\tilde{H}_{b}$\;
			\If{$\mathscr{L}_{a} + \mathscr{L}_{b} < \mathscr{L}^*$}{
				$f\left[node\right].split \leftarrow \left(q,j\right) $\;
				$\mathscr{L}^* \leftarrow \mathscr{L}_{a} + \mathscr{L}_{b}$\;
			}
		}
	}
	\If(\Comment{recursive split}){$\mathscr{L}_0 < \mathscr{L}^*$}{
		$\tilde{g}_a, \tilde{h}_a, \tilde{g}_b, \tilde{h}_b \leftarrow \tilde{g}, \tilde{h}, f\left[node\right].split$ \;
		$x_a, y_a, x_b, y_b\leftarrow x,y, f\left[node\right].split$\;
		$\textbf{\texttt{fit-tree}}\left(x_a,y_a,\tilde{g}_a, \tilde{h}_a, f, node_a\right)$\;
		$\textbf{\texttt{fit-tree}}\left(x_b,y_b,\tilde{g}_b, \tilde{h}_b, f, node_b\right)$\;
	}
	\Else(\Comment{compute best response}){
		\quad $f\left[node\right].r_{opt} \leftarrow r \left((\tilde{H}+\Lambda)^{-1}\tilde{G}\right)$ 
	}
	\caption{$\textbf{\texttt{fit-tree}}$}\label{alg:fit_tree}
\end{algorithm}

\section{Multivariate Regularization}\label{sec:regularizations}
In this section, we introduce some of the most relevant loss functions and multivariate responses that can be modelled through the proposed MBT.

\subsection{Covariance structure and Smoothing}\label{sec:fourier_th}
Generally speaking, imposing a learning bias on the covariance structure of the target can be beneficial for any machine learning algorithm. The most known example of this is linear regression fitting under generalized least squares; in this case, the estimated covariance matrix of the errors $\hat{\Omega}$ is used to penalize the model's errors differently. This can be readily integrated using a linear response function (as explained in section \ref{sec:hierarchical_th}). Under a constant model response, $r(w_l) = w_l$, the covariance structure of the data can be taken into account by means of the quadratic regularization matrix $\Lambda$. For example, we can impose a given smoothness of the response using a filtering approach \citep{Kim2009} such as an Hodrick-Prescott filter \citep{DeJong2016}, punishing the discrete second-order derivative of $r$. This can be obtained setting $\Lambda = \lambda D^TD$ where $D \in \mathds{R}^{(n_t-2) \times n_t}$ is the second-order difference matrix:
\begin{linenomath*}\begin{equation}
	D=\left[\begin{array}{ccccccc}
	1 & -2 & 1 & & & \\
	& & \ddots & \ddots & \ddots & \\
	& & 1 & -2 & 1 & \\
	& & & 1 & -2 & 1
	\end{array}\right]
	\end{equation}\end{linenomath*}  
Since under constant response $h = \mathds{I}_{n_t}$ where $\mathds{I}_{n_t}$ is the identity matrix of dimension $n_t$, we have:
\begin{linenomath*}\begin{equation}
	\Lambda + H = \lambda D^TD + n_l \mathds{I}_{n_t}
	\end{equation}\end{linenomath*}
where $n_l$ is the number of observations in the current leaf. The previous expression can be replaced in \eqref{eq:optimal_response_mgb} and \eqref{eq:optimal_loss_mgb} to retrieve the optimal response and loss of MBT, respectively.

Imposing a condition on the derivative smoothness of the response can be seen as a way to perform signal denoising. If the Hodrick-Prescott filter is applied in a forecasting task, the approach becomes similar to denoising the time series with an a priori smoothing. However, imposing smoothness of the forecasted signal gives the regressor a chance to predict statistically significant peaks, that would have been smoothed out in the pre-processing phase. 

A second approach to induce prediction regularization is through smoothing via basis function \citep{Ramsay2009}. As recently proposed in \cite{Oreshkin2019} in the context of forecasting with neural networks, we can couple a Fourier expansion with the MBT algorithm. We define the response as $r = Pw$ where $P \in \mathds{R}^{n_t \times 2 n_k}$, is a projection matrix onto sine and cosine function space with $n_k$ different wavenumbers:
\begin{linenomath*}\begin{equation}
	P = \left[\left\{\left(cos\left(k \frac{2\pi t}{n_t}\right), sin\left(k \frac{2\pi t}{n_t}\right)\right)_{t=1}^{n_t}\right\} \right]_{k \in \mathcal{K}}
	\end{equation}\end{linenomath*}
where $\mathcal{K}$ is the set of considered wave numbers. Under L2 loss, the $i_{th}$ component of the loss function gradient and Hessian can be written as:
\begin{align}
\tilde{g}_i &= -P^T g_{i}  \\
\tilde{h}_i &= P^T P = \mathds{I}_{n_r}
\end{align}
where the last equality holds due to the fact that P is orthonormal. Under these conditions \eqref{eq:optimal_response_mgb} then becomes:
\begin{linenomath*}\begin{equation}\label{eq:fourier_opt_w}
	w_l^* = -\left(\Lambda + n_l\mathds{I}_{n_r}\right)^{-1} P^T G
	\end{equation}\end{linenomath*}
where $G = \sum_{i\in \mathcal{I}_l} g_{i}$, and \eqref{eq:optimal_loss_mgb} becomes:
\begin{linenomath*}\begin{equation}\label{eq:fourier_opt_loss}
	\mathscr{L}_l^* =  G^T P \left(\Lambda + n_l\mathds{I}_{n_r}\right)^{-1} P^T G =  G^T \left(\Lambda + n_l\mathds{I}_{n_r}\right)^{-1} G
	\end{equation}\end{linenomath*}
where the last equality holds again for the orthonormality of $P$, and $\Lambda$ being diagonal.
\subsection{Latent variables and hierarchical forecasting}\label{sec:hierarchical_th}
In several applications, we are interested in responses that are linear combinations of a fixed matrix $S \in \mathds{R}^{n_t \times n_r}$. That is, $S$ is kept constant through leaves and boosting rounds, while the response $r = S w_l$ can change conditionally to the observations. This procedure restricts the response to lie in the span of $S$. When the dimensionality of $w_l$ is smaller than the response ($n_w<n_t$), $w_l$ can be seen as latent variables generating the full response. Latent variables are usually used to induce regularization in regression \citep{Izenman1975}. Loosely speaking, it is easy to see that all the (conditional) information which is needed to generate $y \in \mathds{R}^{N\times n_t}$ is already present in $x \in \mathds{R}^{N\times n_f}$ if $y^T = Cx^T + \varepsilon$, where $C \in \mathds{R}^{n_t \times n_f}$ is constant, and $\varepsilon \in \mathds{R}^{N\times n_t}$ is the realization of a Gaussian random variable. A notable application of this approach is what is known as hierarchical forecasting; this method tries to reconcile previously produced point forecasts for hierarchically structured signals, by ensuring that the corrected forecasts are consistent under addition. In brief, every time we want to predict a set of base or bottom signals and their groupings (aggregations), we face the problem of making the forecasts aggregate-consistent. Consistency under aggregation is not guaranteed if we separately forecast the bottom time series, call them  $y_b \in \mathds{R}^{N \times n_b}$, and their groupings generated by aggregations $y_u \in \mathds{R}^{N \times n_u}$. The simplest method to have a set of aggregate-consistent forecasts is apply the so-called bottom-up approach, in which only the bottom time series are forecasted, and the forecasts for the aggregated time series are generated by summing them up according to the grouping. This naive approach has been shown to be in general worse than generating forecasts for the aggregated time series by optimally combine the bottom forecasts, which is the concept behind hierarchical forecasting.
Denoting the whole set of original forecasts as $\hat{y} = \left[y_u^T ,y_b^T \right]^T \in \mathds{R}^{N \times n_t}$, where $n_t = n_b + n_u$ and $n_b$ and $n_u$ are the number of the bottom and upper time series, hierarchical forecasting consists in finding a set of corrected bottom forecasts, $\tilde{y}_b$, which minimize the overall forecast error and such that the following equation holds:
\begin{linenomath*}\begin{equation}\label{eq:aggregate_consistent}
	\tilde{y}^T = S \tilde{y}_b^T
	\end{equation}\end{linenomath*}
where $\tilde{y}$ are the corrected signals for the whole hierarchy and $S \in \mathds{R}^{n\times n_b}$ is a summation matrix. An example of a three-level summation matrix is the following:
\begin{linenomath*}\begin{equation}
	S=\left[\begin{array}{cccc}
	1 & 1 & 1 & 1 \\
	1 & 1 & 0 & 0 \\
	0 & 0 & 1 & 1\\
	& \mathds{I}_4 & & 
	\end{array}\right]
	\end{equation}\end{linenomath*}  

In \cite{Hyndman2011}, the authors used ordinary least squares regression to reconcile the forecasts in the hierarchy. Elaborating on this approach, in \cite{Wickramasuriya2017a} and in \cite{Wickramasuriya2018}, the authors proposed a trace minimization method (called minT) in which the covariance matrix of the forecasters' error is estimated to perform a weighted least squares regression. The basic idea exploited in all the aforementioned works is that forecasts can be reconciled solving a generalized least squares problem with error covariance matrix $\hat{\Omega} \in \mathds{R}^{n_t\times n_t}$:
\begin{linenomath*}\begin{equation}\label{eq:gls}
	\tilde{y}^T = S \ \argmin{z} \Vert \hat{y}^T-S z^T\Vert_{\Omega^{-1}}^2
	\end{equation}\end{linenomath*}
which has an analytical solution. Imposing the first derivative to zero, we get:
\begin{linenomath*}\begin{equation}\label{eq:gls_solution}
	\tilde{y}^T = S\left( S^T\Omega^{\dag} S \right)^{-1}S^T\Omega^{\dag}\hat{y}^T 
	\end{equation}\end{linenomath*}
where $\dag$ denotes the pseudo-inverse, since $\Omega$ is typically near-singular. Different hierarchical reconciliation methods basically differ in the choice and estimation of the error covariance matrix $\Omega$.  We can see how \eqref{eq:gls_solution} exploits only information of the originally forecasted signals, and of $\Omega$. The latter is usually estimated using forecast errors from a training set (or from all the available observations), and as such, can be considered invariant. We propose to use a MBT to estimate the reconciled signals starting from $\hat{y}$. This is easily obtained by setting the response to $r=Sw$. Since $S$ is fixed, following the same reasoning of the Fourier decomposition approach introduced in \ref{sec:fourier_th}, equations \eqref{eq:fourier_opt_w} and \eqref{eq:fourier_opt_loss} become:
\begin{align}\label{eq:hf_opt_w}
w_l^* &= -(\Lambda + n_l S^T S )^{-1}S^T G\\
\mathscr{L}_l^* &= G^T S (\Lambda + n_l S^T S )^{-1} S^T G
\end{align}  
The advantage of using a MBT over computing $\tilde{y}$ is that we can use additional features to build the trees. We propose to fit the MBT on the residual between the observed signals and the bottom-up reconciliation, $y - \hat{y}_{b} S^T$, such that the final reconciled time series can be written as: 
\begin{linenomath*}\begin{equation}
	\tilde{y}_{mbt} = f\left(\{ (\hat{y_i},\epsilon_i,x_{t,i})_{i=1}^N\}\right) + \hat{y}_{b} S^T
	\end{equation}\end{linenomath*}
where $\epsilon_i = \hat{y}_{t-1}-y_{t-1} \in \mathds{R}^{N \times n_t} $ contains the forecast error at the timestep prior to the reconciliation and $x_t$ contains categorical encoding of the weekday and the day-hour. Including $\epsilon_i$ in the tree features gives a possibility to the MBT to trust the forecast of the $i_{th}$ predictor, based on its current performances. 

%
%

\subsection{Quantile loss and its relaxations}\label{sec:quantile_th}
\begin{figure}
	\centering
	\includegraphics[width=1\linewidth]{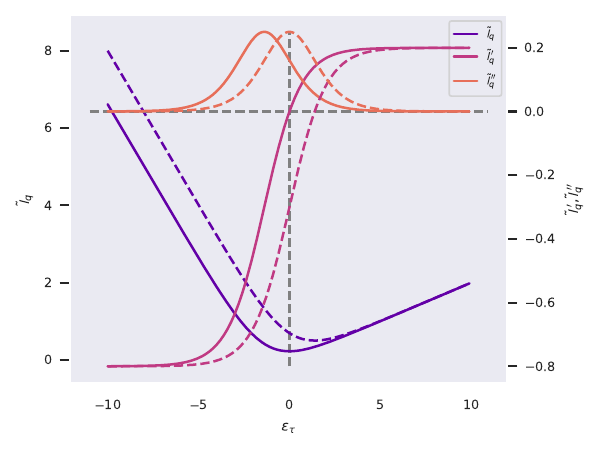}
	\caption{Continuous lines: smoothed quantile loss $\tilde{l}_q(\epsilon_{\tau_i})$ and its first and second derivatives for $\tau_i=0.2$. Dashed lines: the same functions, for  $s=0$.}
	\label{fig:quantile_loss}
\end{figure}

Quantile estimation in the context of boosting is usually achieved by minimizing the so-called quantile loss function, defined as:
\begin{linenomath*}\begin{equation}\label{eq:quantile_loss}
	l_q(\epsilon_{\tau_i}) = \left(\tau_i -\mathds{1}_{\epsilon_{\tau_i}<0}\right) \epsilon_{\tau_i}
	\end{equation}\end{linenomath*}
where $\epsilon_{\tau_i} = y - \hat{q}_{\tau_i}$ is the distance between the observations and the predictions for the $\tau_i$ quantile. It can be shown that the expectation of (30) is minimized when $\hat{q}_{\tau_i}$ is the $\tau_i$ quantile of $F_Y$, $q_{Y}(\tau_i) = F^{-1}_y(\tau_i) = \text{inf} \ \{y:F_Y(y)>\tau_i\}$, for any cdf $F_Y$.
The quantile loss \eqref{eq:quantile_loss} is linear and asymmetric, with an undefined derivative at $\epsilon_{\tau_i} = 0$ and constant 0 Hessian. These characteristics make it hard to exploit the second-order approximation strategy. Indeed, relying only on the first-order approximation reduces the boosting strategy to fitting a classifier on the sign of $\epsilon_{\tau_i}$ at each iteration $k$. Some popular boosted tree packages, like XGBoost, relax the loss function \eqref{eq:quantile_loss} considering a constant second derivative equal to 1. This has the practical effect of fitting the $k_{th}$ model $f_k$ to the leaf-average binary response $\mathds{I}_{\epsilon_{\tau_i}>0}$. We propose a further relaxation of the problem, approximating the discontinuous gradient of the quantile loss function with a smooth function. The idea of smoothing the quantile loss for fitting boosted models was already introduced in \cite{Zheng2012}, where the authors propose to use the cumulative density function of the Gaussian distribution ($\textrm{erf}(\epsilon_{\tau_i})$) as a smoothed version of the gradient of \eqref{eq:quantile_loss}. The rationale behind smoothing $l_q$ is that the MBT will have additional information on how far the observations are from the predicted quantile, which can help in building the tree. In this paper, we decided to use the (scaled and shifted) inverse logit function as a smoothed version of $\tilde{l}_q$ derivative, due to its relation with logistic regression literature and the AdaBoost algorithm (see appendix \ref{ax:connections_with_adaboost}). This choice can be explained by the fact that the distance of the predicted $\tau_i$ quantile from the observation, i.e. $\epsilon_{\tau_i}$, is interpreted as the re-weighted log-odds of the condition $\epsilon_{\tau_i}>0$.
That is, if we describe y as the observation drawn from the random variable $Y(x)$, given the prediction $\hat{q}_{\tau_i}(x)$, we assume:
\begin{linenomath*}\begin{equation}\label{eq:eps_dist}
	\epsilon_{\tau_i}(x) = y - \hat{q}_{\tau_i}(x) = log\left(\frac{(1-\tau_i) F_{Y\vert x} }{  \tau_i(1 - F_{Y\vert x} )}\right)    
	\end{equation}\end{linenomath*}
where $F_{Y\vert x}$ is the conditional cdf of $Y$. Inverting \eqref{eq:eps_dist} we obtain:
\begin{linenomath*}\begin{equation}
	F_{Y\vert x}= \frac{e^{\epsilon_{\tau_i}(x)+s}}{1+e^{\epsilon_{\tau_i}(x)+s}}
	\end{equation}\end{linenomath*}
where $s$ is $logit(\tau_i)$. It can be easily verified that $F_{Y\vert x}=\tau_i$ when $\epsilon_{\tau_i} = 0$. In other words, we are implicitly assuming that the estimated quantile $\hat{q}_{\tau_i}$ is the correct one, under the hypothesis of $Y$ having a logistic pdf: 
\begin{linenomath*}\begin{equation}\label{eq:quantile_hessian}
	h_{i,i} = dF_{Y\vert x} = \frac{e^{\epsilon_{\tau_i}+s}}{(1+e^{\epsilon_{\tau_i}+s})^2}
	\end{equation}\end{linenomath*}
where $h_{i,i}$ is the $i_{th}$ diagonal element of the Hessian of the loss function.
We can now define the smoothed derivative of $l_q(\epsilon_{\tau_i})$ as:
\begin{linenomath*}\begin{equation}\label{eq:quantile_grad}
	-g_k =  -\frac{\partial \tilde{l}_q(\epsilon_{\tau_i})}{\partial{f_k(x)}} = \frac{\partial \tilde{l}_q(\epsilon_{\tau_i})}{\partial{\epsilon_{\tau_i}}} = F_{Y\vert x} -1 + \tau_i
	\end{equation}\end{linenomath*}
and we can now see that its second derivative is equal to the probability density function $\eqref{eq:quantile_hessian}$. 
Since $-1 + \tau_i$ is a constant, and at each iteration we fit $f_k(x)$ on $-g_k$, we can interpret the boosting procedure under the smoothed loss function as an iterative fitting on the probability  $p_{\{Y<\hat{q}_{\tau_i}\vert x \}}$. We can see how the hypothesis on the distribution of the residuals we made in \eqref{eq:eps_dist}, and especially the $\tau_i$ re-weighting, has the effect of shifting $\tilde{l}_q$ such that its minimum is located in $\epsilon_{\tau_i} = 0$. The effect of changing $s$ can be seen in Fig. \ref{fig:quantile_loss}.
As shown in Fig. \ref{fig:quantile_loss}, $\tilde{l}_q(\epsilon_{\tau_i})$ and its derivatives are now smooth functions, thus we can apply the same second-order approximation for fitting the multivariate tree, presented in section \ref{sec:gradient_boosting}.

\paragraph{Linear-quadratic quantile loss function}\mbox{}\\
Smoothing $l_q(\epsilon_{\tau_i})$ has two main drawbacks. First, we cannot guarantee anymore its minimizer being the $\tau_i$ quantile of $F_{Y \vert x}$, independently from its distribution. In fact, any minimizer of $\mathds{E} \left( l_q(\epsilon_{\tau_i})\right)$ must zero its derivative, and this is true for any distribution $F_{Y\vert x}$ only if the derivative is independent from $F_{Y \vert x}$. The second drawback is that, as we try to mitigate the first effect by narrowing the pdf, the objective function becomes closer to the original quantile loss, turning the regression problem again in a classification one.
Here we introduce a linear-quadratic quantile loss function which is consistent for any target pdf. We exploit the learning peculiarities of trees to approximate $F_{l, Y\vert x}$ in each leaf with the empirical one, $\hat{F}_{l, Y\vert x}$, and craft a smooth objective function whose minimizer is the empirical quantile of the $\hat{F}_{l, Y\vert x}$.
%
%

\begin{theorem}\label{th:1}
	Given a sample population $\mathcal{D} = \{(x_i,y_i)_{i=1}^N\}$, $\epsilon_{\tau, i} = y_i - \hat{q}_{\tau}(x_i)$ being the distance between the $i_{th}$ observed target and its predicted $\tau$ quantile, $k$ being a constant, the following loss function:
	\begin{linenomath*}\begin{equation}
		\begin{aligned}\label{eq:squared_loss}
		l_{qs,i}(\epsilon, \hat{q}_{\tau}(x_i),\tau) &= \left( (\tau-1)\epsilon_{\tau, i} +\frac{k\epsilon_{\tau, i}^2}{2\bar{\epsilon}_{\tau, l}}\right)\mathds{1}_{y<\hat{q}_{\tau}} \\ 
		& + \left(\tau \epsilon_{\tau, i} +\frac{k\epsilon_{\tau, i}^2}{2\bar{\epsilon}_{\tau,r}}\right)\mathds{1}_{y\geq \hat{q}_{\tau}} -\epsilon_{\tau, i} 2\frac{k}{N}
		\end{aligned}
		\end{equation}\end{linenomath*}
	
	where $\bar{\epsilon}_{\tau,l} = \sum_{i \in \mathds{I}_l} \epsilon_{\tau, i}$, $\mathds{I}_l=\{i: y_i < \hat{q}_{\tau}(x)\}$, $\bar{\epsilon}_{\tau,r}=\sum_{i \in \mathds{I}_r} \epsilon_{\tau, i}$,$\mathds{Is}_r=\{i: y_i \geq \hat{q}_{\tau}(x)\}$, is minimized by the empirical quantile of $Y = {(y_i)_{i=1}^N}$.
\end{theorem}
The proof is reported in appendix A. The diagonal entries of the Hessian are then:
\begin{align}\label{eq:squared_loss_hessian}
h_{i,i} = \frac{\partial^2 l_{qs,i}}{\partial \epsilon_{\tau, i}^2} &= \left(\frac{k}{\bar{\epsilon}_{\tau,l}}\right)\mathds{1}_{y\leq \hat{q}_{\tau}}\\ &\nonumber + \left(\frac{k}{\bar{\epsilon}_{\tau,r}}\right)\mathds{1}_{y\geq \hat{q}_{\tau}}
\end{align}

As recently introduced in the LightGBM implementation, we also consider the case of refitting the leaf responses $w_l$. After fitting the weak learner $f_k$ using one of the approximated previously introduced losses, we replace $w_l$ with the exact minimizers of \eqref{eq:quantile_loss}, given the identified tree regions. That is, for each $\tau_i$:
\begin{linenomath*}\begin{equation}
	w_{l,i} = \hat{F}_{Y\vert x_l}^{-1}(\epsilon_{k,l,\tau_i})
	\end{equation}\end{linenomath*}
where $\epsilon_{k,l,\tau_i}$ is the error at iteration $k$ for the current leaf and quantile $\tau_i$, while $\hat{F}_{Y\vert x_l}^{-1}$ is the inverse of the empirical conditional cdf of the current leaf.

\subsection{Data driven control}\label{sec:vsc_th}
Standard control methods rely on a model of the controlled system, which is usually identified through system identification techniques \citep{ljungSystemIdentification1998a}. One standard description of the controlled system is the so called linear state-space representation, which in its discrete time-invariant form is described by:
\begin{align}\label{eq:lin_sys}
\xi_{t+1} &= A \xi_t + B u_t + G w_t \\ 
\gamma_{t+1} &= C\xi_t + D u_t + H w_t + v_t \label{eq:lin_sys_2} 
\end{align}
where $\xi_t \in \mathds{R}^{n_s}$ is the vector of system states, $\gamma_t \in \mathds{R}^{n_o}$ is the vector of measured system's outputs, $u_t \in \mathds{R}^{n_o}$ is the vector of system's controlled inputs and $w_t \in \mathds{R}^{n_s}$ and $v_t \in \mathds{R}^{n_o}$ are two vector of (usually) uncorrelated Gaussian disturbances, taking into account discrepancy between the system's model and the real one and measurement noise, respectively. Model \eqref{eq:lin_sys}-\eqref{eq:lin_sys_2} is then used to optimally control the target system, usually coupling it with feedback controllers or with model predictive control (MPC) \citep{morariModelPredictiveControl1988a}. Data-driven control (DDC) has been introduced in the last years as a way to overcome identification issues in MPC. For many systems of interest, a single linear system could not provide enough accuracy, while increasing the number of states or switching to a non-linear system can introduce identification issues and increase the computational time of the controller. The authors in \cite{Jain2017,Smarra2018} introduce the idea of fitting a tree $f(x,\theta)$, which responses are linearized dynamics of the controllable system. If the features used for growing the tree do not include control actions and system states, the linear dynamics identified in the leaves can be regarded as independent from the system and thus be directly used for control. Overcoming identifiability issues for control application is of great practical interest, and as such DDC gained popularity in the last year \citep{datadriven}. Here we propose to apply MBTs to increase the accuracy of the identified linear models, with respect to the one identifiable with a single tree. In this case, the weak learner $f(x,x_{lr},\theta)$ requires two sets of features: the one used to grow the tree and choose the best split $x \in \mathds{R}^{N \times n_f}$, and the one used to fit the linear model in each leaf $x_{lr} \in \mathds{R}^{N \times n_{lf}}$. Note that, due to the additive nature of boosting, the final model will still be a linear system in the tree's inputs. In this case, the second-order approximation is not helpful to reduce the calculation effort, because it corresponds to the exact solution of a linear system. We have, in fact:
\begin{linenomath*}\begin{equation}\label{eq:lin_reg_w}
	w_l^* = -\left(\Lambda + x_{lr,l}^T x_{lr,l} \right)^{-1} x_{lr,l}^T g_{l}
	\end{equation}\end{linenomath*}
where $x_{lr,l} \in \mathds{R}^{n_l \times n_f}$ is the feature matrix in the current leaf, and $g_l \in \mathds{R}^{n_l \times n_t}$ is the gradient matrix in the current leaf.

\begin{table}[]
	\caption{List of combinations of loss and response functions, with their gradients and Hessians. First row: constant response with second derivative regularization. In rows $2,3,4$ responses are linear functions of: nonlinear basis function, constant summation matrix, feature space. Last row: different quantile loss approximations with a constant response.}
	\label{tab:loss_list}
	\begin{tabular}{@{}lllll@{}}
		\toprule
		Section                 & $\mathscr{L}$                   &  $r$               & $\tilde{G}$               & $\tilde{H} + \Lambda$                \\ \midrule
		\multicolumn{1}{|l|}{\ref{sec:fourier_th}} & \multicolumn{1}{l|}{L2} & \multicolumn{1}{l|}{$w$} & \multicolumn{1}{l|}{$\epsilon$} & \multicolumn{1}{l|}{$n_l \mathds{I}_{n_r} + \lambda D^T D $} \\ \midrule
		\multicolumn{1}{|l|}{\ref{sec:fourier_th}}  & \multicolumn{1}{l|}{L2}  & \multicolumn{1}{l|}{$Pw$}  & \multicolumn{1}{l|}{$P^T \epsilon $}  & \multicolumn{1}{l|}{$n_l \mathds{I}_{n_r} + \Lambda$}  \\ \midrule  
		\multicolumn{1}{|l|}{\ref{sec:hierarchical_th}}  & \multicolumn{1}{l|}{L2}  & \multicolumn{1}{l|}{$Sw$}  & \multicolumn{1}{l|}{$S^T \epsilon $}  & \multicolumn{1}{l|}{$n_l S^TS+ \Lambda$}  \\ \midrule  
		\multicolumn{1}{|l|}{\ref{sec:vsc_th}}& \multicolumn{1}{l|}{L2}  & \multicolumn{1}{l|}{$x_{lr, l}w$}  & \multicolumn{1}{l|}{$\epsilon x_{lr, l}^T$}  & \multicolumn{1}{l|}{$x_{lr, l}^T x_{lr, l} + \Lambda$}  \\ \midrule
		\multicolumn{1}{|l|}{\ref{sec:quantile_th}}  & \multicolumn{1}{l|}{$l_q(\epsilon_{\tau})$}  & \multicolumn{1}{l|}{$w$}  & \multicolumn{1}{l|}{\eqref{eq:eps_dist} / \eqref{eq:squared_loss}}  & \multicolumn{1}{l|}{\eqref{eq:quantile_hessian} / \eqref{eq:squared_loss_hessian}}  \\ \midrule
	\end{tabular}
\end{table}

\subsection{Consistency}
The additive nature of boosting guarantees consistency in the properties encoded in the weak learners, if they are invariant under summation. 
The two smoothing approaches presented in section \ref{sec:fourier_th} show different levels of consistency under boosting. For the Hodrick-Perscott filter, at each iteration, a curve with penalized second derivative is added in each leaf, such that the final curve is still smooth. However, if we compute the quadratic loss for the final response, $\left(\sum_{k=1}^{n_i} w_{l,k} \right)^TD^TD \left(\sum_{k=1}^{n_i} w_{l,k} \right)$ could be higher than the same loss from a single weak learner. This means that the final level of smoothness could depend on the number of fitting rounds $n_i$.  For the Fourier expansion case, the final response will be a summation over Fourier coefficients in the chosen wave numbers $k \in \mathcal{K}$, which means the final signal will be a superposition of columns of $P$. This means that the Fourier decomposition property of identifying a signal composed only by harmonics with $\mathcal{K}$ wave numbers is fully retained.
The single fitted responses in section \ref{sec:hierarchical_th}  respects the hierarchical relationship encoded in $S$, that is $r_k = Sw_{l,k}$. Since $S$ is constant through leaves and boosting rounds, also the final prediction retain this property, since $r = \sum_{k=1}^{n_i} S w_{l,k} = S \sum_{k=1}^{n_i} w_{l,k}$. 

Quantile losses of section \ref{sec:quantile_th} do not generate strictly consistent responses. This is because the quantiles corrections identified at each iteration $k$ are not jointly constrained. However, we will see in section \ref{sec:quantile} that in the case of the refitting strategy, consistency is respected in practice, presenting very few quantile crossing instances.

Finally, the prediction of MBT with linear responses of the feature space, like the one in section \ref{sec:vsc_th}, is consistently linear in $x$, being a superposition of linear functions.

\subsection{Numerical Methods}
Table \ref{tab:loss_list} summarizes the forms of the loss gradients and Hessian for the different combinations of losses and responses introduced in the previous section. In particular, the last column contains the expression that needs to be inverted when computing the optimal response $w_l^*$ and approximated loss function. Inverting $\tilde{H} + \Lambda$ requires most of the computational time of the algorithm. Thus it is important to try to simplify or speed up this computation. In \cite{Zhang2019}, the authors present an upper bound for the optimal response and loss in the case of a constant response and when the matrix $\tilde{H} + \Lambda$ is diagonally dominant. 
Here we show how to accelerate the exact computation of $(\tilde{H} + \Lambda)^{-1}$ for three of the cases in table \ref{tab:loss_list}. 
We can see how the first two cases require to invert a constant (through leaves and boosts) matrix, plus the identity matrix multiplied by the number of elements in the current leaf, $n_l$. Called $A\in\mathds{R}^{k, k}$ this matrix, this inversion can be reduced to a matrix multiplication in the form $	(A + n\mathds{I}_k)^{-1} = Q^T \tilde{L} Q$ where $\tilde{L} \in \mathds{R}^{k,k}$ is diagonal with $\tilde{L}_{i,i} = 1/(\lambda_{A_i} + n)$ and $\lambda_{A_i}$ is the ith eigenvector of $A$, thanks to lemma \eqref{lemma} reported in appendix \ref{ax:th_2}, along with its proof. Since in our case $A$ is constant, its eigenvalues, $Q$ and its inverse can be computed only once for the entire fitting process. The only variable part is $n$, which in our case corresponds to the number of observations in the current leaf. This only affects the diagonal entries of $\tilde{L}$, while all the other quantities remain unchanged. For the third case of table \ref{tab:loss_list}, we have to invert $n_l S^TS + \Lambda$. Once again, the only non-constant term is $n_l$. If the quadratic regularization term $\Lambda$ is a multiple of the identity matrix (as is typically assumed), this can be written as $n_l S^TS + n \mathds{I}$, and we can use the following corollary of lemma \eqref{lemma}:
\\ \\
{\bf Corollary}{\label{corollary}
	\it Given a symmetric invertible matrix $A \in \mathds{R}^{k \times k}$, $(m A + n\mathds{I}_k)^{-1}$  can be computed as:
	\begin{linenomath*}\begin{equation}\label{eq:matrix_inv_2}
		(mA + n\mathds{I}_k)^{-1} = Q L Q^{-1} / m							
		\end{equation}\end{linenomath*}
	where $L \in \mathds{R}^{k,k}$ is diagonal with $L_{i,i} = 1/(\lambda_i + n / m)$, and $\lambda_i$ and $Q$ as defined in \eqref{lemma}.
}
\\ \\
the latter corollary follows from lemma \eqref{lemma} proof, noting that $mA + n\mathds{I}_{k} = m(A + n\mathds{I}_{k}/m)$. 

\section{Numerical results}\label{sec:numerical_results}
\begin{figure}
	\centering
	\includegraphics[width=0.5\linewidth]{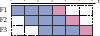}
	\caption{Example of CV for 3 folds. Rows and columns indicate different folds and different times, respectively. Blue: training sets. Violet: test sets.}
	\label{fig:cv}
\end{figure}

In this section, we present numerical results of the responses and loss functions introduced in section \ref{sec:regularizations}. For all the datasets, we obtained the results using k-fold cross-validation (CV). Since all the applications deal with temporal data, we adopted sliding-window cross-validation. An example of training and testing splits under this cross-validation is shown in Fig. \ref{fig:cv}, in the case of 3 folds. In all the experiments the hyperparameters were fixed to the following values, in order to guarantee a fair comparison with the LightGBM regressors. For all the experiments, we kept the LightGBM's number of iterations fixed to 100 and a learning rate of 0.1, as for the MBT models. Table \ref{tab:pars} shows the most important parameters for the different experiments carried out in the paper. The $min_l$ parameter specifies the minimum number of observation in one leaf. We set a minimum number of 10 observations per feature for the VSC experiment, since in this case we need to solve a linear regression in each leaf. At the same time, we lower the value of $\lambda$ to 0.01 in this case, since we didn't expect presence of noise in the simulated dataset.

\begin{table}[h!]\label{tab:pars}
	\centering
	\caption{Valuse for the most important hyperparameters, as a function of the numerical experiment (and the corresponding section in brackets).}
	\begin{tabular}{l|l|l|l|l|}
		\cline{2-5}
		& $n_{boost}$ & learning rate & $min_l$ & $\lambda$ \\ \hline
		\multicolumn{1}{|l|}{Fourier (\ref{sec:fourier})}     & 100   & 0.1           & 300       & 1        \\ \hline
		\multicolumn{1}{|l|}{Hierarhical (\ref{sec:hierarchical})} & 100   & 0.1           & 400       & 1        \\ \hline
		\multicolumn{1}{|l|}{Quantiles (\ref{sec:quantile})}   & 100   & 0.1           & 300       & 1        \\ \hline
		\multicolumn{1}{|l|}{VSC (\ref{sec:vsc})}         & 100   & 0.1           & 10 $n_f$ & 0.01     \\ \hline
	\end{tabular}
\end{table}

\subsection{Forecasting via Fourier decomposition}\label{sec:fourier}
We applied the Fourier-based MBT introduced in \ref{sec:fourier_th} to two public datasets, available at \citep{public_dataset} and \citep{competitionsM4methods2022}. The first one consists of about 1 year of electrical load measurements of secondary substations and cabinets located in a low voltage distribution grid, and additional numerical weather predictions for the temperature and the irradiance. The signals have a sampling frequency of 10 minutes. In total, 31 time series are provided, showing hierarchical relationships, that is, 7 time series are the algebraic summation of specific subgroups. Called $P_i$ the power measurement of the $i_{th}$ time series, we aim at forecasting the day-ahead signal (144 steps), given historical values of the power, the numerical weather predictions of temperature and irradiance, and time-related covariates:
\begin{linenomath*}\begin{equation}
	\hat{P}_{i, t} = f(P_{t-j},x_t, x_{f, t+z})
	\end{equation}\end{linenomath*}  
where $x_t$ contains categorical encodings of the weekday and the day-hour, $x_{f, t+z}$ contains the numerical weather predictions of temperature and irradiance at time $t+z$ and $z, j \in [1,144]$, meaning that we pass to the forecaster all the numerical weather predictions and an history of the power signal of 24 hours. We compared the MBT with two baselines using LightGBM and two different multi step-ahead strategies \citep{BenTaieb2012}. The first one mimics a multiple-input multiple-output approach (MIMO). This is obtained, similarly to what is done in \cite{Sampathirao2014a} with support vector machines, by adding an auxiliary feature $x_{c,i}$ to the dataset, which represents a categorical encoding of the step ahead to which $y_i$ corresponds.  The second one adopts a multiple-input single-output (MISO) approach: 144 different models are trained, each of them predicting a given step ahead. This strategy has the advantage of increasing the final forecaster flexibility, at the price of disregarding time correlations in the predictions.    

\begin{figure}
	\centering
	\includegraphics[width=1\linewidth]{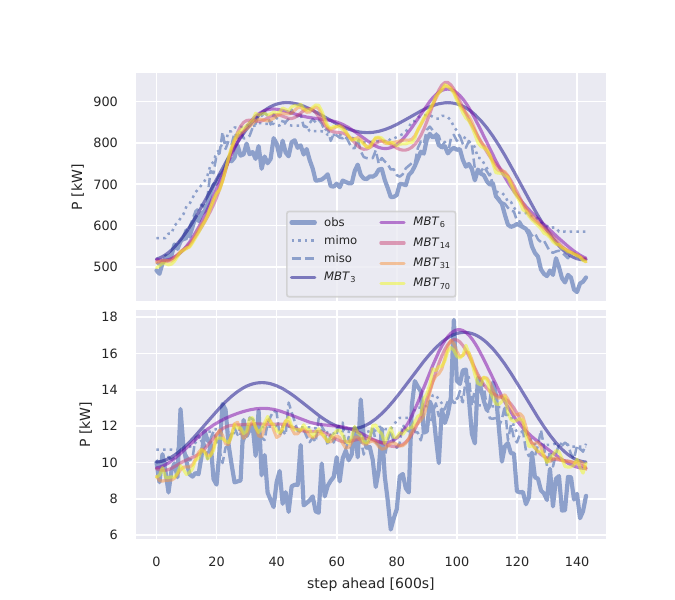}
	\caption{Example of forecasting via Fourier decomposition on the aggregated time series (top) and on time series belonging to the lower aggregation level (bottom). Thick blue line represents the ground truth, while the dotted and dashed lines represents the mimo and miso benchmarks. The other lines are forecasts obtained with MBT, the color indicating an increasing number of considered frequencies, from darker to lighter.}
	\label{fig:fourier_smoothing_example}
\end{figure}

\begin{figure}
	\centering
	\includegraphics[width=1\linewidth]{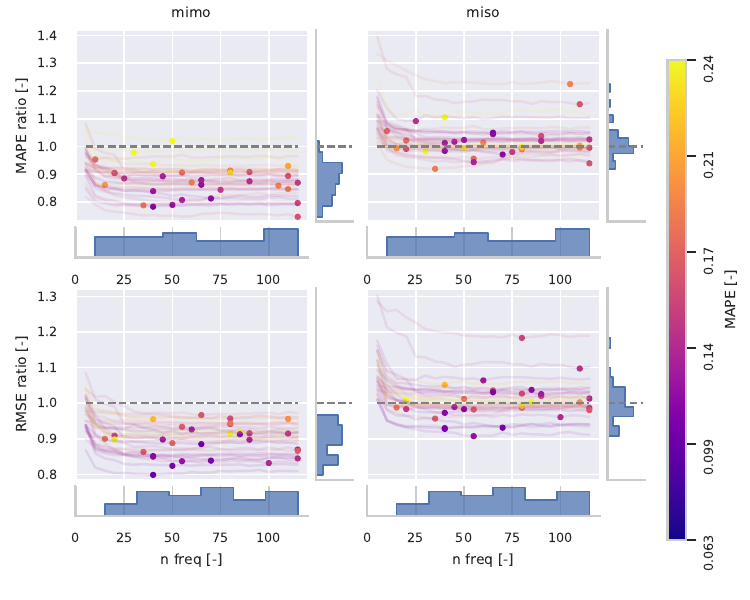}
	\caption{RMSE and MAPE, mediated over CV folds and step ahead, for a different number of harmonics fitted by the forecaster based on Fourier decomposition. In the first column, the KPIs are normalized with the KPIs of the MIMO forecaster, while in the second one they are normalized with the KPIs of MISO strategy. Colours in the first and second column refer to the MAPE of the MIMO and MISO strategy, respectively. Histograms show the distributions of the best KPIs as a function of the number of fitted harmonics, marked as a dot for each case.}
	\label{fig:fourier_smoothing}
\end{figure}

An example of 24 hours ahead Fourier forecasting using an increasing number of harmonics is shown in Fig. \ref{fig:fourier_smoothing_example}. The top panel shows the aggregated time series, while the second panel shows one of the bottom (more variable) time series. It can be seen how increasing the number of harmonics (from dark to light colours) increases the flexibility of the forecaster while keeping potential useful time correlations. However, in this case, the targets present a degree of correlation which depends on the hour of the day. In the top panel of Fig. \ref{fig:fourier_smoothing_example} it can be seen how the target is strongly correlated in the early morning and during evening hours, while correlation is less obvious in during the day. This pattern is recurrent in all the days of the dataset. 
To see the effect of the number of harmonics on the accuracy of the MBT, we retrieve the forecasts for all the 31 time series using a 3 fold CV, for an increasing number of wavenumbers. This investigation is reported in Fig. \ref{fig:fourier_smoothing}, where the CV fold-mediated and normalized RMSE and MAPE are reported. The first column uses the values of the RMSE and MAPE from the MIMO strategy benchmark for the normalization of the results, while the second one normalizes the MBT key performance indicators (KPIs) with the one obtained with the MISO strategy. Dots highlights the best normalized performance for the various time series, while colours represent the MAPE obtained with MIMO (first column) and MISO (second column) strategies.
We can see how the MBT is strictly better than the MIMO strategy in terms of RMSE, for almost all the number of harmonics, while achieving better results in terms of MAPE for all but one case. Despite the lack of inter-temporal information, the MISO strategy performs better than the other two on average. The MBT provides higher accuracy for 14 time series in terms of RMSE and for 11 in terms of MAPE. However, no evident correlation with respect to the MISO strategy MAPE (line colour) is observed.        

In all the cases, we can observe an initial improvement of performances with respect to increasing wavenumber. Results show that the minimum of the KPIs lies in what looks like a plateau for all the considered cases, as the wavenumber increases. This means that while considering more harmonics than the one highlighted by the dots, the accuracy does not increase or decrease significantly. This suggests that including a priori information on the smoothness (and time correlation structure) of the curve doesn't seem to be particularly helpful for this dataset. This is possibly due to the fact that the available features are already very informative for the prediction of the power signal, and further imposing a regularization on the temporal shape of the prediction doesn't help the regression. 

\begin{figure}
	\centering
	\includegraphics[width=1\linewidth]{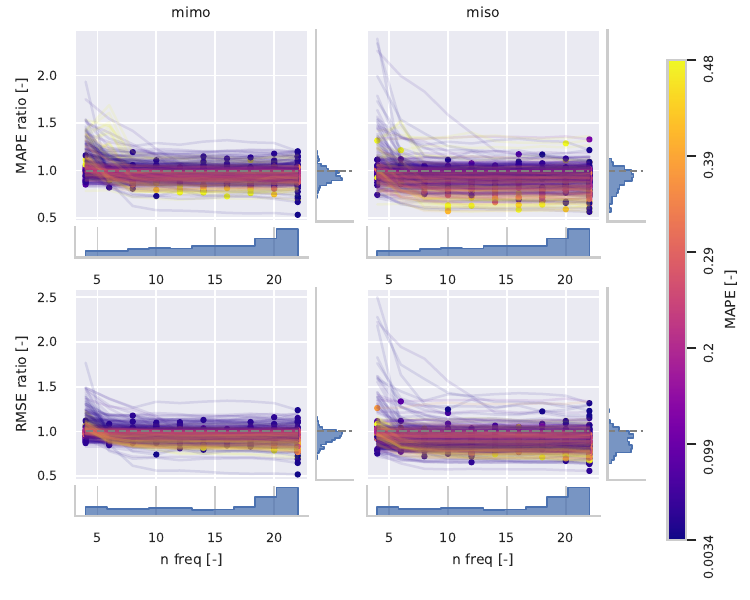}
	\caption{Same of figure \ref{fig:fourier_smoothing}, but for the M4 hourly dataset.}
	\label{fig:m4_fourier_smoothing}
\end{figure}

\begin{figure}
	\centering
	\includegraphics[width=1\linewidth]{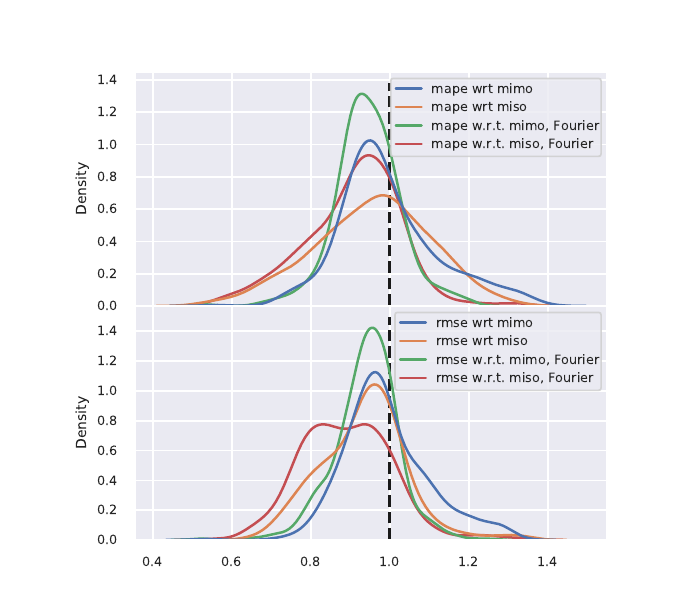}
	\caption{Comparison of the MBT model and the MBT model using Fourier regularization w.r.t. the normalized MAPE and RMSE of the MIMO and MISO strategies, in terms of distributions for the 414 time series of dataset \citep{competitionsM4methods2022}.}
	\label{fig:m4_regression}
\end{figure}

In order to test this hypothesis, we applied the same methods on the hourly dataset of the M4 competition \citep{competitionsM4methods2022}, which do not have associate exogenous features. We discarded those time series containing missing values and tested the method on a total of 414 signals. In this case the predictions at time $t$ for the $i_{th}$ time series are given by $\hat{y}_{i, t} = f(y_{i, t-j})$ where $j\in [1, 48]$, meaning that we passed a two days history of the signal to predict the next day ahead. The results w.r.t. the MIMO and MISO strategy are shown in figure \ref{fig:m4_fourier_smoothing}. We can see how the distribution of the best performing number of harmonics is skewed towards high numbers, as opposed to the much more uniform distributions of figure \ref{fig:fourier_smoothing}. At the same time, for most of the time series MBT obtains a better performance in both MAPE and RMSE, as can be seen from the vertical distributions of figure \ref{fig:m4_fourier_smoothing}. To actually see if the increase of performance is due to the Fourier regularization, in figure \ref{fig:m4_regression} we compared the MBT model and the MBT model using Fourier regularization w.r.t. the normalized MAPE and RMSE of the MIMO and MISO strategies, in terms of distributions for the 414 time series. Switching from the base MBT model to the Fourier regularized one causes the distributions of the MAPE and the RMSE to shift towards smaller values, both when normalized with the MISO and the MIMO results.

\subsection{Hierarchical forecasting}\label{sec:hierarchical}
\begin{figure}
	\centering
	\includegraphics[width=1\linewidth]{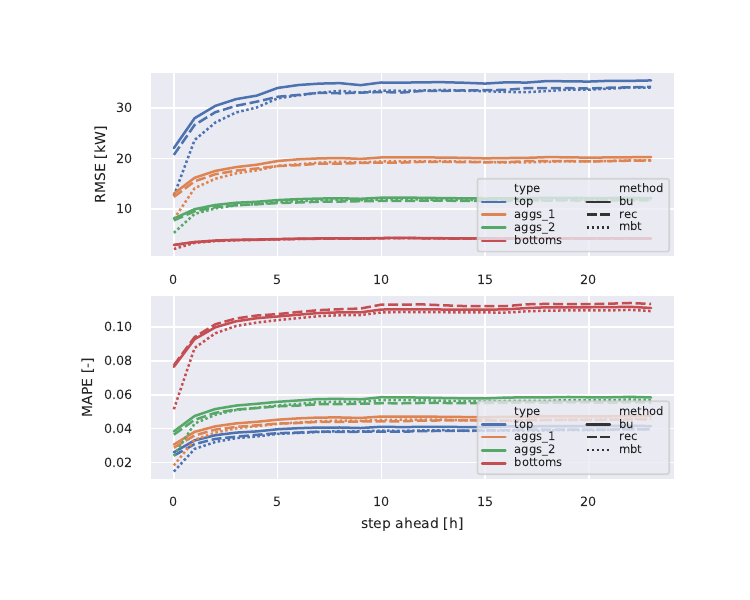}
	\caption{RMSE as a function of the step ahead, grouped by hierarchical levels, mediated over the CV folds, for different reconciliation techniques. Blue, orange, green and red lines refer to the overall aggregated profile, the first and second level of aggregates, and the bottom time series, respectively. Continuous lines: bottom-up reconciliation. Dashed lines: reconciled forecasts using the shrink strategy and glasso covariance estimation. Dotted lines: MBT with history of reconciliation errors.}
	\label{fig:reconciliation_rmse}
\end{figure}

Using the same dataset of the previous section, we obtained the baseline 24 hours ahead forecasts for all the 31 time series, using 3 fold CV. 
In this dataset we have 3 aggregation levels, so that the summation matrix $S$ can be written as:
\begin{linenomath*}\begin{equation}
	S = \left[
	\begin{matrix}
	&\mathds{1}_{n_b}\\
	&I_{2} \otimes \mathds{1}_{n_b/2}\\
	&I_{4} \otimes \mathds{1}_{n_b/4}\\
	&I_{n_b}
	\end{matrix}
	\right] 
	\end{equation}\end{linenomath*}
where $\mathds{1}_{n_b}$ is the unit row vector with the size equal to the number of bottom-level time series, in this case, $n_b = 24$ and $\otimes$ is the Kronecker product. The forecasts are then reconciled using the minT strategy \citep{Wickramasuriya2018}, coupled with the graphical Lasso approach \citep{Friedman2008} for the error covariance estimation and a bottom-up strategy. The latter consist in retrieving consistent forecast summing up bottom level forecasts. Formally, we obtain the set of reconciled forecasts as $\tilde{y} = S \hat{y}_b$. We then compare the results with a MBT using information about the forecast error of the previous timestep, as described in section \ref{sec:hierarchical_th}. The results as a function of the step ahead, and divided by aggregation groups, are presented in Fig. \ref{fig:reconciliation_rmse}. We can see how the additional information that MBT can exploit significantly decrease the forecast error for the first hours ahead. On the other hand, the advantage over standard reconciliation approaches vanishes with the increase of the step-ahead. Since the MBT requires substantially more computational time, an effective strategy would be to fit this model only for the initial steps ahead and then switch to the standard reconciliation strategy.  

\subsection{Boosted voltage sensitivity coefficients}\label{sec:vsc}

While DDC has been mainly applied to the control of heating systems, here we propose an application for the control in the electrical distribution grid. 
When performing optimal power flow, a distribution system operator (DSO) must take into account the nonlinear power flow equation, which includes the nonlinear relation:
\begin{linenomath*}\begin{equation}
	S = V \odot I^*
	\end{equation}\end{linenomath*}
where $S$, $V$ and $I$ are the vectors of complex powers, voltages and currents in the buses of the network, $^*$ denotes the complex conjugate and $\odot$ the Hadamard product. Different relaxations of power flow equation exist \citep{Molzahn2017}. Usually, either the knowledge of phasors' angles (e.g. DC approximation) or the knowledge of the lines' parameters and topology (e.g. the DistFlow model) are required inputs to this approximation. However, this information is not always available. For example, the network topology of the low-voltage grid, where most residential users are located, is usually unknown or difficult to access. In the absence of network topology, one could opt for an approximate formulation of the power flow, whose parameters can be estimated using smart meter data. One of these formulations consists of the first-order linearization of the power flow equations. The linear coefficients of this formulation are known as the voltage sensitivity coefficients (VSCs):  
\begin{linenomath*}\begin{equation}
	\begin{aligned}
	k_{i,j}^p &= \frac{\partial \vert V_j \vert}{\partial P_k} 
	&k_{i,j}^q &= \frac{\partial \vert V_j \vert}{\partial Q_k}
	\end{aligned}
	\end{equation}\end{linenomath*} 
where $P$ and $Q$ are the active and reactive power, respectively, and $k_{i,j}^p$,$k_{i,j}^q$ are the sensitivity coefficients between node $i$ and node $j$. The analytical expression of voltage sensitivity coefficients, and an efficient method to compute them based on the state of the grid and admittance matrix, is provided in \cite{Christakou2013}. 
In \cite{Mugnier2016}, it has been shown that the voltage sensitivity coefficients can be estimated by least-squares regression of the time derivatives of voltage magnitudes, $P$ and $Q$. We follow their approach to find sets of VSCs, conditional to the state of the grid. However, knowing the latter would require to know all the voltages of the buses' grid. As discussed in section \ref{sec:vsc_th}, we aim at building the MBT without using the state of the system, we use the power measurements at the point of common coupling (PCC) with the medium voltage grid as a proxy for the state of the grid.

\begin{figure}
	\centering
	\includegraphics[width=0.6\linewidth]{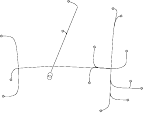}
	\caption{Schematics of the simulated low voltage grid for the VSC computation. Grey circles indicates single households, while the double circle indicate the power transformer.}
	\label{fig:lic_grid}
\end{figure}

In order to compare the approach in \cite{Mugnier2016} with the MBT one, we simulated 3 months of data for a low voltage grid located in Switzerland. Fig. \ref{fig:lic_grid} show the topology of the grid and the QP buses locations. This information, along with parameters for the grid's cables, were retrieved from the local DSO. Power profiles of uncontrollable loads were generated with the LoadProfileGenerator \citep{Pflugradt2013}; power profiles of photovoltaic roof-mounted power plants were obtained through the PVlib python library \citep{Stein2012}, while the electrical loads due to heat pumps was retrieved simulating domestic heating systems and buildings thermal dynamics, modelling them starting from building's metadata. The grid was then simulated with OpenDSS \citep{Dugan2012}, and the 3 phases voltages, power and currents retrieved for all the QP nodes of the grid, with a 1 minute sampling time.

The results were then obtained by applying a 10 fold CV. As an additional comparison, we considered Ridge regression for the VSCs. Since $x_{lr}$ and $y$ both have a high number of dimensions, quadratic regularization could help in finding a better solution. The regularization coefficient for the Ridge regression was found using an inner CV for each fold. The dataset for the linear regression was $\mathcal{D} = \{(x_{lr,i},y_i)_{i=1}^N\}$ is the same for all the three models, where $x_{lr} \in \mathds{R}^{N\times6n}$ contains the discrete-time derivatives of P and Q values for the 3 phases of all the buses, while $y \in \mathds{R}^{N\times3n}$ contains the time derivatives of the voltages. The MBT was built using $x \in \mathds{R}^{N\times3}$, which contains $P_{PCC}$, which is the power measured at the PCC (the double circle in Fig.  \ref{fig:lic_grid}), the hour of the day and the weekday. In this case, the tree growth is not independent from the control action, since the power at PCC includes the power of controlled appliances in the grid. Under these conditions, the MBT can still be applied to build an oracle for checking voltage violations, using a "proxy-Lagrangian" formulation of the optimization problem \citep{Cotter2019}. However, this results in a more complex formulation, being the constraints non-convex. We compare this solution to one in which the MBT is only fitted using meteorological variables, i.e. the ambient temperature $T_a$ and the solar irradiance $G_{irr}$, the hour of the day and the weekday. In this case the identified leaves are independent from system state and control actions, and as such the MBT can be employed in standard convex optimization. 

\begin{figure}
	\centering
	\includegraphics[width=1\linewidth]{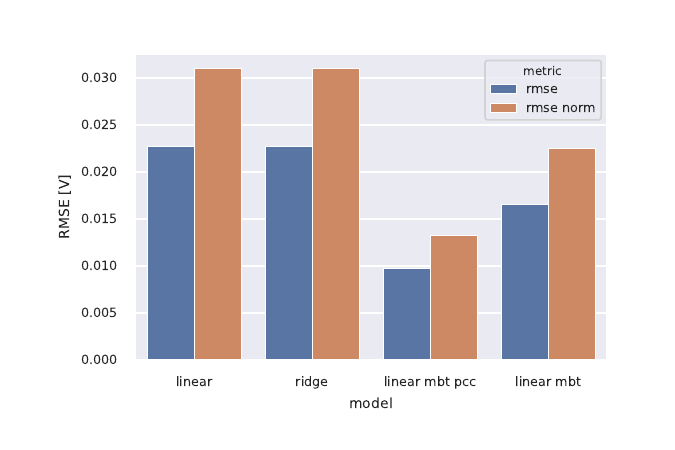}
	\caption{RMSE for different regressors, mediated over the CV folds and nodes. The red columns show the RMSE normalized with the predictions using the sample mean values.  }
	\label{fig:vsc_rmse}
\end{figure}

\begin{figure}
	\centering
	\includegraphics[width=1\linewidth]{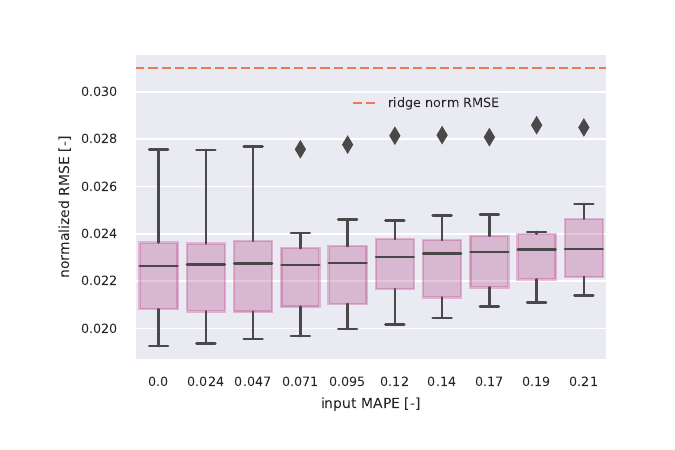}
	\caption{Boxplots for the 10 fold CV of the normalized RMSE of the MBT predictor for increasing levels of noise in the tree inputs, in terms of MAPE.}
	\label{fig:vsc_noise}
\end{figure}

Fig. \ref{fig:vsc_rmse} shows results in terms of mean RMSE over folds and grids' nodes, and of normalized RMSE. The normalization of the latter is done with the mean RMSE obtained with a constant prediction of the voltage. This is a meaningful normalization because in Europe voltage signals in low voltage networks have a nominal value of 230V, and usually they do not deviate more than the 10$\%$. Ridge regularization slightly increase the accuracy, while the MBT does it significantly. As expected, the MBT using power at PCC is more accurate with respect to its counterpart using only disturbances for the growth of the trees. This means that the power at the PCC is a better proxy for the state of the electrical grid than the meteorological variables. However, since these models are meant to be used in control applications, the models must be accurate for the whole decision horizon (typically 24 hours ahead for demand-side management applications). Since the first two models are constant, they do not need any further investigation. On the other hand, the final MBT model depends on the features used to build the tree, $x$. In the following we restrict the analysis to the MBT fitted on the meteorological variables; the one fitted on P at PCC shows a very similar behavior. Indeed, we only need to investigate the accuracy degradation with respect to the forecasted $T_a$ and $G_{irr}$, since the other two variables in $x$ are deterministic. We thus applied increasing levels of multiplicative noise from a (3 $\sigma$) truncated Gaussian distribution to $T_a$ and $G_{irr}$, to mimic accuracy degradation in its forecasts, and retrieved the MBT normalized RMSE on the CV folds. The results are shown in Fig. \ref{fig:vsc_noise} in terms of increasing MAPE on the forecasted $T_a$ and $G_{irr}$ signals,  as box plots containing the 10 CV folds measurements. We can conclude that the degradation of the MBT is negligible up to a MAPE of $21 \%$, which corresponds to very bad forecasts for this kind of applications.

\subsection{Quantile prediction}\label{sec:quantile}
We tested the different quantile loss relaxations and fitting strategies presented in section \ref{sec:quantile_th} on the aggregated power profile of the hourly-resampled dataset \citep{public_dataset}. In particular, we seek to retrieve the quantile predictions tensor $\hat{q}_{\tau_i} \in \mathds{R}^{N \times n_t \times n_q}$ where $\tau_i \in \mathscr{T}$, $\mathscr{T}$ is a set of $n_q = 11$ equispaced quantiles and $n_t = 24$. For all the methods, we kept the same features and target matrix $x$ and $y$ as specified in section \ref{sec:fourier}. The benchmark to which we compare the MBT-based solutions are 24 sets of $n_q$ LightGBMs, that is, we fitted a different LightGBM for each combination of step-ahead and quantile. Other three models are then compared: the MBT using the smoothed version of quantile loss $\tilde{l}_q$, defined by its gradient \eqref{eq:quantile_grad} and Hessian \eqref{eq:quantile_hessian}; the same model with quantile refitting, as explained in section \ref{sec:quantile_th}; the model using the linear-quadratic quantile loss defined by its gradient  \eqref{eq:squared_loss} and Hessian \eqref{eq:squared_loss_hessian}, with quantile refitting.  

Quality of quantile forecasts is harder to assess compared to point forecasts since different desirable properties of the forecasted prediction interval must be evaluated. For this comparison we relied on 4 KPIs. The first one is the time average of the quantile loss \eqref{eq:quantile_loss}, $\bar{l}_q = \sum_{t=1}^T l_q(\epsilon_{\tau_i,t})$.
The second one is the quantile score $Qs(\hat{q}_{\tau_i},y)$, which is a proper scoring rule \citep{Gneiting2007a,Golestaneh2016a,Bentzien2014}, and it's defined as the expected quantile loss \eqref{eq:quantile_loss}:
\begin{align}
Qs &= \int_0^1 \bar{l}_q(\epsilon_{\tau_i}) \mathrm{d} \tau_i \simeq \sum_{\tau_i \in \mathscr{T}} \bar{l}_q(\epsilon_{\tau_i}) \mathrm{d} \tau_i
\end{align}
where $\hat{q}_{\tau_i}$ is the predicted $\tau_i$-quantile, while $y$ is the observed ground truth. This score is strictly connected with the continuous rank probability score \citep{Gneiting2007a}, both being total variation measurements between the forecasted pdf and a Heaviside distribution centered on the observation y. For these scores, lower values indicate better performances. The third KPI is based on the reliability \citep{Pinson2010}, which is the average number of times the observed signal was actually below the predicted $\tau$ quantile.

\begin{linenomath*}\begin{equation}
	r_{\tau_i} = \frac{1}{N} \sum_{j=1}^{N} \mathds{1}_{\{y_j < \hat{q}_{\tau_i,j}\}}  
	\end{equation}\end{linenomath*}
When plotted against $\mathscr{T}$, the perfect reliability aligns with the bisector of the first quadrant. Because all the models provided highly reliable quantiles, to ease the comparison of the performance, we defined the following KPI:
\begin{linenomath*}\begin{equation}
	Rs = \vert r_{\tau_i}(F_b)-\tau \vert - \vert r_{\tau_i}(F_m)-\tau \vert
	\end{equation}\end{linenomath*}
that is, the difference of absolute deviation from the perfect reliability, between a benchmark forecasting model $F_b$ and the considered one, $F_m$.
The last KPI is the mean crossing of the quantiles. We define it as:
\begin{linenomath*}\begin{equation}
	\overline{\chi} =  \frac{1}{n_q} \sum_{i=1}^{n_q} \frac{1}{N} \sum_{j=1}^{N} \mathds{1}_{\hat{q}_{\tau_i}>\hat{q}_{\tau_{i+1}}}
	\end{equation}\end{linenomath*}
that is, the average over quantiles of the mean number of times $\hat{q}_{\tau_i}$ violates the monotonicity of $\hat{F}(Y)$. 

\begin{figure}
	\centering
	\includegraphics[width=1\linewidth]{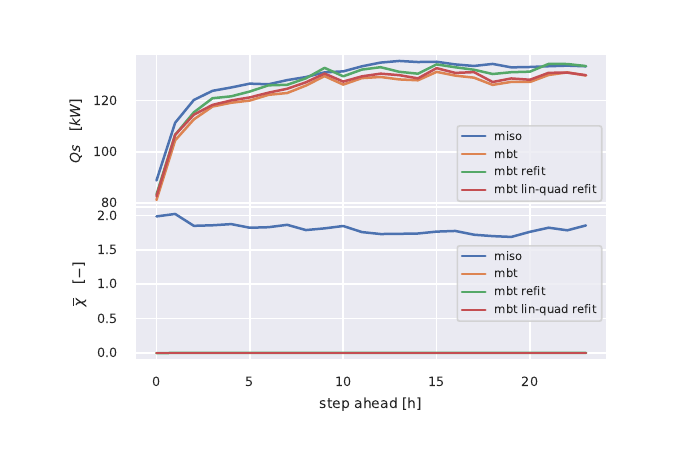}
	\caption{CRPS and mean quantile crossings, as a function of step-ahead, mediated over the CV folds. The refitted MBT with logistic and quadratic losses show similar performances with the MISO strategy while achieving a significantly lower number of crossings.}
	\label{fig:Qscore}
\end{figure}

\begin{figure}
	\centering
	\includegraphics[width=1\linewidth]{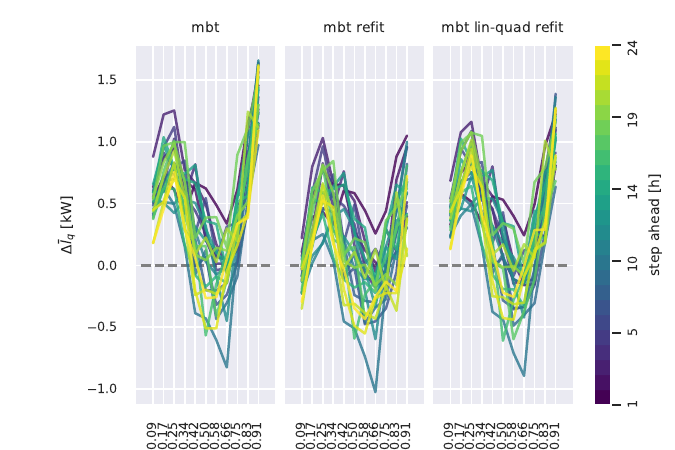}
	\caption{Differences between $\bar{l}_q$ for different models, with respect to the benchmark case, as a function of $\tau_i$ and the step ahead (line colour, from blue to yellow). Lines above the grey dashed line denotes better performances.}
	\label{fig:QS_diff}
\end{figure}

\begin{figure}
	\centering
	\includegraphics[width=1\linewidth]{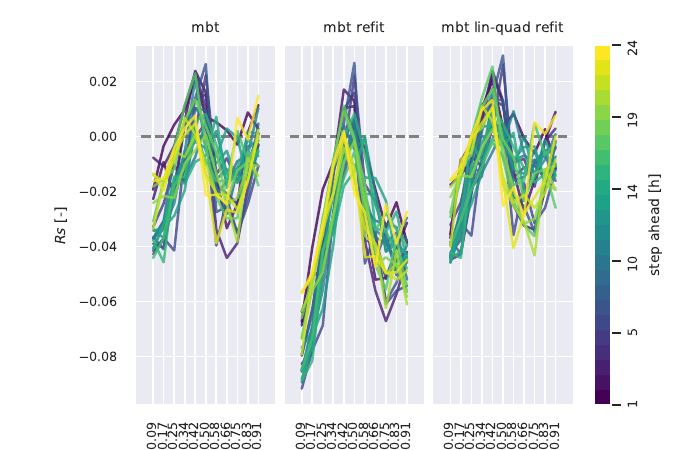}
	\caption{$Rs$ for different models, with respect to the benchmark case, as a function of $\tau_i$ and the step ahead (line colour, from blue to yellow). }
	\label{fig:reliability_diff_tilted}
\end{figure}

In Fig. \ref{fig:QS_diff} we compare the quantile loss $l_q(\epsilon_{\tau_i})$ as a function of $\tau_i$ and the step ahead. To ease the comparison, we plot the differences between the $l_q(\epsilon_{\tau_i})$ of the benchmark and the other models. The original quantile loss plots can be found in the appendix \ref{ax:additional_figs}. All the MBT models are consistently better at modelling the tails of the distribution, while applying refitting to the linear quantile loss function doesn't show any improvement. In terms of reliability, Fig. \ref{fig:reliability_diff_tilted} shows how both the base model and the one using the lin-quantile loss have similar reliability with respect to the benchmark. The first panel of Fig. \ref{fig:Qscore} shows the quantile score $Qs$ as a function of step-ahead for the four different models. All the MBT based models show a $Qs$ score lower than the benchmark, the base MBT model and the one using the linear-quadratic formulation being strictly better for all the steps ahead. The second panel shows $\overline{\chi}$ for increasing steps ahead. It is evident how using different BTs for different quantiles leads the benchmark model to inconsistent results. The quantile crossing is negligible for all the MBT based models when compared to the benchmark. 

\section{Conclusions}
In this paper, we have presented a multivariate boosted tree algorithm, fitted using the same second-order Taylor expansion used by LightGBM and XGboost. The algorithm allows to arbitrarily regularize the predictions, through the use of multivariate penalization and basis functions. 
We have shown how, for a relevant class of applications, the Hessian inversion required for fitting the underlying tree models can be reduced to a matrix multiplication, making the algorithm computationally appealing. Unlike its univariate counterpart, the MBT is particularly useful when properties like smoothness, consistency and functional relations are required. We have shown this through numerical examples on four different tasks, namely: time series forecasting, hierarchical reconciliation, data-driven control and quantile forecasting. While including a priori regularization on the smoothness of a forecasted time series doesn't seem to increase accuracy against univariate BTs with a MISO strategy, for the other presented applications, where some consistency is explicitly required, the algorithm showed clear advantages. We conclude by noting that the presented MBT algorithm only used histogram-based split search since we did not make use of very large datasets in our experiments. Computational time can be readily reduced through the adoption of numerical techniques tailored to tree fitting, such as gradient-based one-side sampling and exclusive feature bundling \citep{Ke2017}.  

\section{Aknowledgemnts}
This  project  is  carried  out  within  the  frame  of  the  Swiss  Centre  for Competence in Energy Research on the Future Swiss Electrical Infrastructure(SCCER-FURIES) with the financial support of the Swiss Innovation Agency (Innosuisse  -  SCCER  program)  and  of  the  Swiss  Federal  Office  of  Energy(project SI/501523)

\appendix
\section{Proof of theorem \ref{th:1}}\label{ax:th_1}
\begin{proof}
	The loss function (35) is minimized in expectation, with respect to the empirical distribution of the target $y$ in $\mathcal{D} = \{(x_i,y_i)_{i=1}^N\}$, if the expectation of its derivative is zeroed by its minimizer:
	\begin{align}
	q^* &= \argmin{q} \mathds{E}_{\mathcal{D}} l_{qs}(x,q,\tau)\\
	q^* & \  s.t. \ \frac{\partial \mathds{E}_{\mathcal{D}} l_{qs}(x,q^*,\tau)}{\partial q} = 0 \label{A2}
	\end{align}
	Keeping the same nomenclature in theorem (1), we retrieve $q^*$ by solving \eqref{A2}. We recall that the derivative of the set membership function $\mathds{1}_{z>0}$ is zero almost everywhere, and by the chain rule, deriving $f(z)\mathds{1}_{z>0}$ results in $\frac{\partial{f(z)}}{\partial z}\mathds{1}_{z>0}$. Since we want the derivative of the expectation over the dataset $\mathcal{D}$, we have
	\begin{align*}
	\frac{\partial \mathds{E}_{\mathcal{D}} l_{qs}}{\partial q} &= \frac{1}{N} \sum_{i=1}^N \bigg[ \left( (\tau-1) +\frac{k\epsilon_{\tau, i}}{\bar{\epsilon}_{\tau, l}}\right)\mathds{1}_{y<\hat{q}_{\tau}} \\
	&+ \left(\tau +\frac{k\epsilon_{\tau, i}}{\bar{\epsilon}_{\tau,r}}\right)\mathds{1}_{y\geq \hat{q}_{\tau}} -2\frac{k}{N} \bigg]
	\end{align*}
	summation over the $N$ elements of the dataset and set membership functions can be turned into partial summations over the sets $\mathds{Is}_l=\{i: y_i \leq \hat{q}_{\tau}(x)\}$ and $\mathds{Is}_r=\{i: y_i \geq \hat{q}_{\tau}(x)\}$:
	\begin{equation}
	\frac{\partial \mathds{E}_{\mathcal{D}} l_{qs}}{\partial q}  =  \frac{1}{N} \bigg[\sum_{i \in \mathds{Is}_l} \left( (\tau-1) +\frac{k\epsilon_{\tau, i}}{\bar{\epsilon}_{\tau, l}}\right) + \sum_{i \in \mathds{Is}_r}\left(\tau +\frac{k\epsilon_{\tau, i}}{\bar{\epsilon}_{\tau,r}}\right) -2k\bigg]
	\end{equation}
	by the definition of $\bar{\epsilon}_{\tau, l}$ and $\bar{\epsilon}_{\tau, r}$, this further simplifies into:
	\begin{equation}
	\frac{\partial \mathds{E}_{\mathcal{D}} l_{qs}}{\partial q}  = \frac{1}{N} \bigg[  n_l (\tau -1) + k + n_r \tau + k -2k\bigg]
	\end{equation}	
	Given that $n_r = N - n_l$, where $n_l$ and $n_r$ are the cardinalities of the $\mathds{Is}_l$ and $\mathds{Is}_r$ sets, respectively, we get:
	\begin{equation}
	\frac{\partial \mathds{E}_{\mathcal{D}} l_{qs}}{\partial q}  = \frac{N\tau  -n_l }{N}
	\end{equation}	
	and finally, zeroing it we get:
	\begin{equation}
	\tau = \frac{n_l}{N}  
	\end{equation}	
	That is, the optimal $q^*$ minimizing $\mathds{E}_{\mathcal{D}} l_{qs}$  must be greater than exactly a fraction of $\tau$ observations of $y$ contained in the dataset $\mathcal{D}$, which is the definition of the empirical $\tau$ quantile. 
\end{proof}
%

\section{Connections with AdaBoost}\label{ax:connections_with_adaboost}
At each iteration, AdaBoost employs an exponential loss function in order to solve a binary classification problem. It can be shown that the minimizer $f_k^*(x)$ of this loss minimizes also the logit loss associated to the classification probabilities \cite{Friedman2000} :  
\begin{linenomath*}\begin{equation}\label{eq:logit}
	f_k^*(x) = \argmin{f_k(x)} l_{A}(y,f_k(x)) =  \ log\left(\frac{P_{\{y=1\vert x\}}}{P_{\{y=-1\vert x \}}}\right)
	\end{equation}\end{linenomath*}
and therefore, inverting this relation, the conditional probability $p(y=1\vert x)$ can be written as:
\begin{linenomath*}\begin{equation}
	p(y=1\vert x) = \frac{e^{f_k^*(x)}}{e^{-f_k^*(x)}+e^{f_k^*(x)}} = \frac{e^{2f_k^*(x)}}{1+e^{2f_k^*(x)}}
	\end{equation}\end{linenomath*}
which means that AdaBoost algorithm can be explained in terms of an additive logistic regression model. 

\section{Matrix inverses}\label{ax:th_2}

\begin{lemma}\label{lemma}
	\it	Given a symmetric invertible matrix $A \in \mathds{R}^{k \times k}$, $(A + n\mathds{I}_k)^{-1}$  can be computed as:
	\begin{equation}\label{eq:matrix_inv}
	(A + n\mathds{I}_k)^{-1} = Q L Q^{-1}
	\end{equation}
	where $L \in \mathds{R}^{k,k}$ is diagonal with $L_{i,i} = 1/(\lambda_i + n)$, $\lambda_i$ is the $i_{th}$ eigenvalue of $A$ and $Q$ is the matrix whose columns are the eigenvectors of $A$.
\end{lemma}

\begin{proof}
	Considering the eigenequation of matrix $A$:
	\begin{equation}
		Ax = \lambda x
		\end{equation}
	and adding a multiple of the identity matrix:
	\begin{linenomath*}\begin{equation}
		(A+n\mathds{I}) x = (\lambda+n)x
		\end{equation}\end{linenomath*}	
	calling $A+n\mathds{I} = \tilde{A}$, this means that $\lambda_{\tilde{A},i} = \lambda_{A,i} + n$, where $\lambda_{A,i}$ denotes the $i_{th}$ eigenvalue of $A$. Since adding a multiple of $\mathds{I}$ to $A$ just influences the magnitude of the vector to which the final matrix is applied, the eigenvectors of $A$ and $A + n\mathds{I}$ are the same. Thus, since $A$ is symmetric and invertible, and its inverse can be obtained as:
	\begin{eqnarray}
	A^{-1} = Q^T L Q
	\end{eqnarray}  
	$\tilde{A}^{-1}$ can be obtained as 
	\begin{eqnarray}
	\tilde{A}^{-1} = Q^T \tilde{L} Q
	\end{eqnarray}  
	where $L \in \mathds{R}^{k,k}$ is diagonal with $L_{i,i} = 1/\lambda_{A_i}$ and $\tilde{L} \in \mathds{R}^{k,k}$ is diagonal with $\tilde{L}_{i,i} = 1/(\lambda_{A_i} + n)$.
\end{proof}
\section{Statistical analysis}
We performed Nemenyi tests \cite{Hollander1999} on the experiments presented in the paper to statistically compare the performances of the different models. The Nemenyi test is a post-hoc pairwise test, which is used to compare a set of $m$ different models on a group of $n$ independent experiments. Firstly, a matrix $R \in \mathrm{R}^{n\times m}$ whose elements $r_{i,j}$ are the ranks for experiment $i$ and model $j$, is obtained. Then, the mean rank for each model is retrieved through column-wise averages of $R$. The performance of two models is identified as significantly different by the Nemenyi test if the corresponding average ranks differ by at least the critical difference:
\begin{equation}\label{eq:conf}
CD=q_{\alpha, m} \sqrt{\frac{m(m+1)}{12 n}}
\end{equation}
where $q_{\alpha}$ is the quantile $\alpha$ of the Studentized range statistic with $m$ samples. We implemented the Nemenyi test in python following the implementation in the \texttt{tsutils} R package \cite{tsutils}.  The Nemenyi test is usually performed after a Friedman’s test, which is a non-parametric analog of variance for a randomized block design; this can be considered as non-parametric version of a one-way ANOVA with repeated measures. More details on the difference and implementation of the two tests can be found in \cite{Dale2006}.
Since we run several experiments through the paper, it is of interest to perform not just one test, but several ones, to assess under which conditions the MBT regressor is better. In the following and in the figures, we refer to the KPI used for building the ranking matrix $R$ as the target variable, and to the parameter or property we have changed through different tests as the independent variable. In the following we present the results of the statistical tests on the experiments presented in the paper. All the preliminary Friedman's tests confuted the null hypothesis that the compared algorithms have the same distribution for the target variable; the only exception was the reliability of the quantile forecasting experiment for the $\alpha=0.5$ quantile, which means that the compared models where considered to be statistically equally reliable.
\begin{figure}[!h]
	\centering
	\includegraphics[width=1\linewidth]{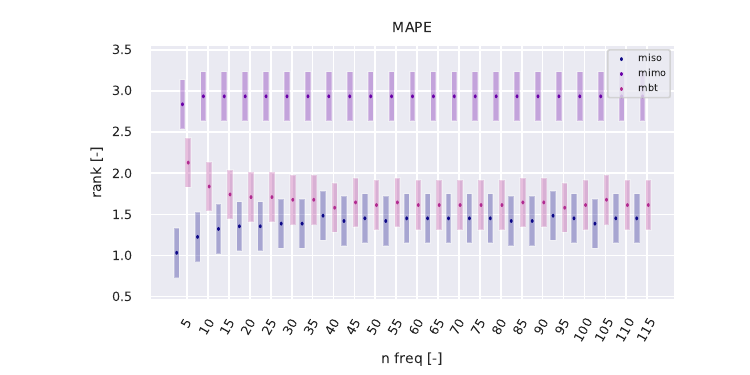}
	\caption{Nemenyi tests for the Fourier forecasting using the \citep{public_dataset} dataset. The tests are grouped by time series, while the independent variable is the number of the harmonics used. The target variable is the MAPE.}
	\label{fig:stats_fourier}
\end{figure}

\begin{figure}[!h]
	\centering
	\includegraphics[width=1\linewidth]{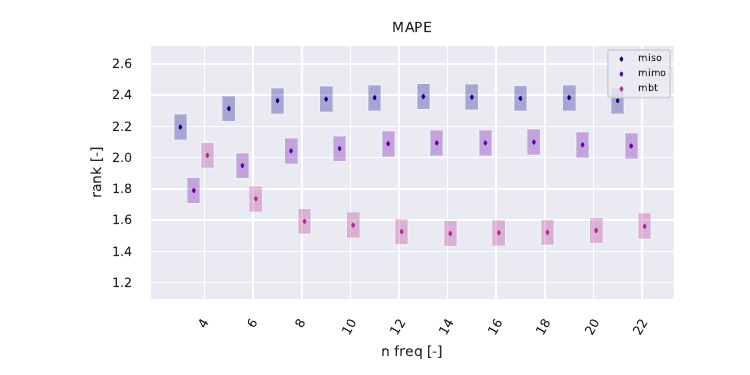}
	\caption{Nemenyi tests for the Fourier forecasting using the \citep{competitionsM4methods2022} dataset. The tests are grouped by time series, while the independent variable is the number of the harmonics used. The target variable is the MAPE.}
	\label{fig:stats_m4}
\end{figure}

\begin{figure}[]
	\centering
	\includegraphics[width=1\linewidth]{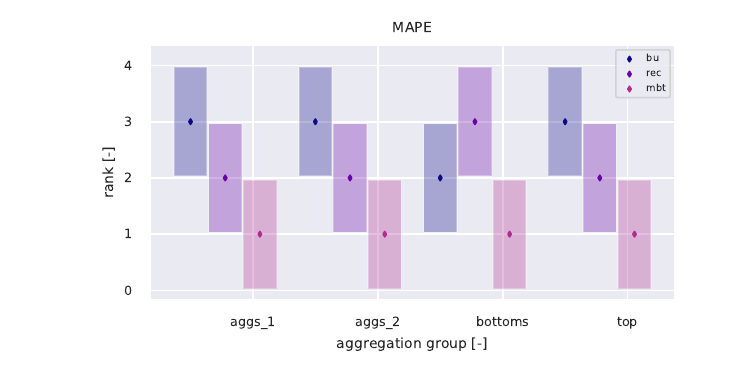}
	\caption{Nemenyi tests for the hierarchical forecasting. The tests are grouped by the first three steps ahead, while the independent variable is the group level. The target variable is the MAPE.}
	\label{fig:stats_hierarchical}
\end{figure}

\begin{figure}[!h]
	\centering
	\includegraphics[width=1\linewidth]{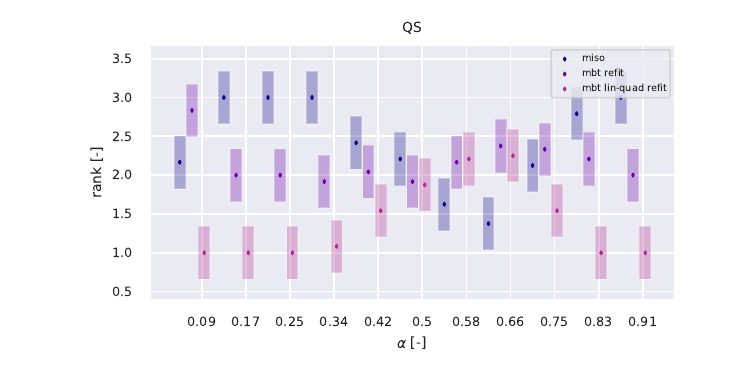}
	\caption{Nemenyi tests for the quantile forecasting. The tests are grouped by step ahead, while the independent variable is the $\alpha$ quantile. The target variable is the quantile score.}
	\label{fig:stats_QS}
\end{figure}

\begin{figure}[!h]
	\centering
	\includegraphics[width=1\linewidth]{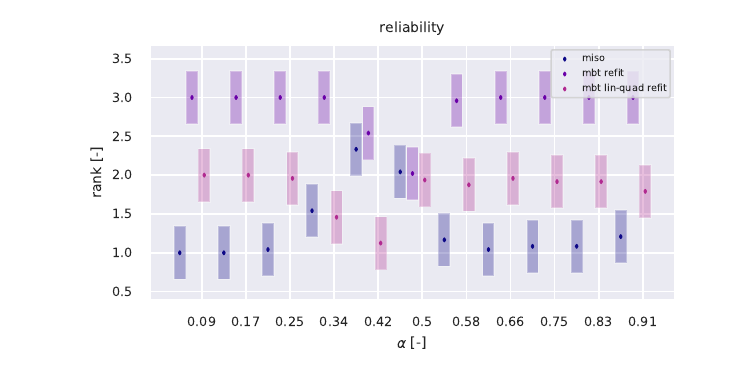}
	\caption{Nemenyi tests for the quantile forecasting. The tests are grouped by step ahead, while the independent variable is the $\alpha$ quantile. The target variable is the reliability.}
	\label{fig:stats_reliability}
\end{figure}

In Fig. \ref{fig:stats_fourier} the column-wise means of the $R$ matrix and the confidence bands obtained through the $CD$ values \eqref{eq:conf} are shown, for the Fourier loss experiments using the dataset \citep{public_dataset}. In this case the population of the reference experiments is composed by the 31 time series, so that in this case we have $n=31$. The target variable is the MAPE of the MIMO, MISO and the MBT models, while the independent variable is the number of harmonics used by the MBT model. The MIMO model is consistently worse than the other two. The MBT model is always better than the MIMO model; while it is worse than the MISO model when using few number of harmonics, its performances gets statistically indistinguishable from the MISO model for a number of harmonics higher than 25. This confirms that inducing smoothness in the multiple step ahead forecasting task doesn't help in reducing the MAPE. Fig. \ref{fig:stats_m4} shows the same analysis but for the dataset \citep{competitionsM4methods2022}. In this case it's clear that the Fourier smoothing helps decreasing the MAPE compared to the MISO and MIMO models.
Fig. \ref{fig:stats_hierarchical} refers to the hierarchical forecast experiments, with the first three steps ahead as population, MAPE as target variable and level of aggregation as independent variable. For each level of aggregation we see that the MBT regressor perform better w.r.t. the bottom up aggregation and the hierarchical reconciliation method. For one aggregation group, the bottom time series, the hierarchical reconciliation worsen the base forecast results, while the MBT regressor consistently performs better also in this case. Fig. \ref{fig:stats_QS} and \ref{fig:stats_reliability} refer to the quantile forecast experiments, with the first three steps ahead as population, level of aggregation as independent variable and quantile score and reliability deviations, defined as $\vert r_{\tau_i}(F_m)-\tau \vert$, as target variable. For the quantile score we see that while the MISO strategy perform better for the central quantiles, both the MBT models (with the normal quantile loss and with the linear-quadratic one) perform better for the extreme quantiles. On the other hand, when considering reliability, the MISO strategy is better for extreme quantiles.

\pagebreak
\FloatBarrier
\section{Additional figures}\label{ax:additional_figs}

\begin{figure}[h]
	\centering
	\includegraphics[width=1\linewidth]{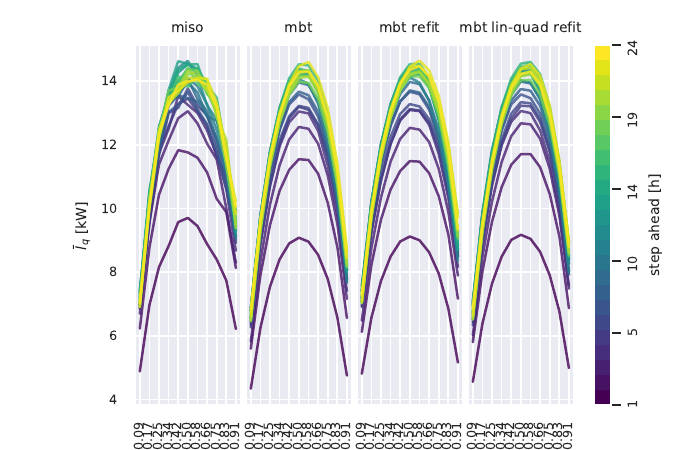}
	\caption{$\bar{l}_q$ for different models, as a function of $\tau_i$ and the step ahead (line color, from blue to yellow).}
	\label{fig:QS}
\end{figure}

\begin{figure}[h]
	\centering
	\includegraphics[width=1\linewidth]{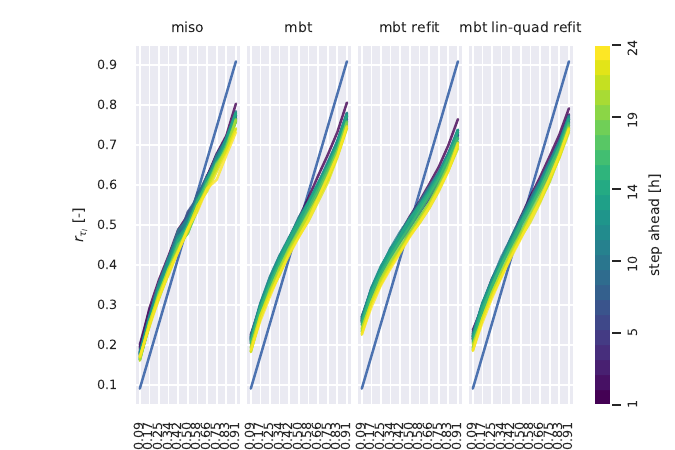}
	\caption{Reliability plots for different models, as a function of $\tau_i$ and the step ahead (line color, from blue to yellow).}
	\label{fig:reliability}
\end{figure}
\FloatBarrier
\pagebreak

\bibliographystyle{elsarticle-num}
\bibliography{MGB}
\nomenclature[A]{MBT}{multivariate boosted tree}
\nomenclature[A]{GBT}{gradient boosted tree}
\nomenclature[A]{pdf}{probability density function}
\nomenclature[A]{cdf}{cumulative density function}
\nomenclature[A]{DDC}{data driven control}
\nomenclature[A]{MPC}{model predictive control}
\nomenclature[A]{VSC}{voltage sensitivity coefficients}
\nomenclature[A]{MIMO}{multiple-input multiple-output}
\nomenclature[A]{MISO}{multiple-input single-output}
\nomenclature[A]{CV}{cross validation}
\nomenclature[A]{RMSE}{root mean square error}
\nomenclature[A]{MAPE}{mean absolute percentage error}
\nomenclature[A]{PCC}{point of common coupling}

\nomenclature[B]{$y$}{target variable matrix}
\nomenclature[B]{$x$}{feature matrix}
\nomenclature[B]{$g_k$, $\tilde{g}_k$}{loss gradient w.r.t. $F_k$, $w_k$}
\nomenclature[B]{$h_k$, $\tilde{h}_k$}{loss Hessian w.r.t. $F_k$, $w_k$}
\nomenclature[B]{$\tilde{G}_k$}{$\sum_{i \in \mathcal{I}_l} \tilde{g}_{i,k}$}
\nomenclature[B]{$\tilde{H}_k$}{$\sum_{i \in \mathcal{I}_l} \tilde{h}_{i,k}$}
\nomenclature[B]{$\varepsilon$}{boosted model training loss}
\nomenclature[B]{$\epsilon$}{prediction error}

\nomenclature[B]{$\mathscr{L}$}{loss function}
\nomenclature[B]{$r$}{response function}
\nomenclature[B]{$F$}{boosted model}
\nomenclature[B]{$f$}{weak learner}
\nomenclature[B]{$\mathds{1}_{x}$}{indicator function on condition $x$}
\nomenclature[B]{$p$}{probability}
\nomenclature[B]{$\hat{y}_b,\tilde{y}_b $}{forecasted and reconciled bottom time series}
\nomenclature[B]{$\hat{y}_u,\tilde{y}_u $}{forecasted and reconciled upper levels time series}
\nomenclature[B]{$\tau$}{quantile level}
\nomenclature[B]{$F_{Y\vert x}$}{cdf of random variable $Y$}
\nomenclature[B]{$\mathds{E}$}{expectation operator}
\nomenclature[B]{$x_{lr}$}{feature matrix for linear response}
\nomenclature[B]{$k_{i,j}^p,k_{i,j}^q$}{VSC for node $i$ w.r.t. node $j$, for power and reactive power }
\nomenclature[B]{$Q_s$}{quantile score}
\nomenclature[B]{$r_{\tau}$}{reliability of quantile $\tau$}
\nomenclature[B]{$\overline{\chi}$}{average number of quantile crossings}

\nomenclature[C]{$N$}{number of observations}
\nomenclature[C]{$n_t$}{targets dimension}
\nomenclature[C]{$n_f$}{features dimension}
\nomenclature[C]{$n_{lf}$}{linear features dimension}
\nomenclature[C]{$n_w$}{leaf's weights dimension}
\nomenclature[C]{$n_l$}{number of leaves}
\nomenclature[C]{$n_b$}{number of bottom time series}
\nomenclature[C]{$n_{qs}$}{number of quantile splits for histogram search}
\nomenclature[C]{$n_q$}{number of predicted quantiles}
\nomenclature[C]{$n_u$}{number of upper level time series}
\nomenclature[C]{$\Theta$}{BT parameters}
\nomenclature[C]{$\theta$}{tree parameters}
\nomenclature[C]{$W$}{response function parameters}
\nomenclature[C]{$w_l$}{leaf-specific parameters}
\nomenclature[C]{$\rho$}{learning rate}
\nomenclature[C]{$\lambda$}{quadratic regularization coefficient}
\nomenclature[C]{$\Lambda$}{quadratic regularization matrix}
\nomenclature[C]{$n_i$}{number of boosting rounds}
\nomenclature[C]{$n_{min}$}{minimum number of observations per leaf}
\nomenclature[C]{$\Omega$}{error covariance matrix}
\nomenclature[C]{$D$}{second order difference matrix}
\nomenclature[C]{$n_k$}{number of wavenumbers}
\nomenclature[C]{$S$}{summation matrix}
\nomenclature[C]{$\mathds{I}_n$}{identity matrix of size $n$}

\nomenclature[D]{$\mathcal{D}$}{dataset}
\nomenclature[D]{$\mathcal{K}$}{set of wavenumbers}
\nomenclature[D]{$\mathcal{I}_l$}{observations in leaf $l$}

\end{document}